\documentclass{ecai}
\usepackage{graphicx}

\usepackage{bm}
\usepackage{bbm}
\usepackage{amssymb,amsfonts}
\usepackage{mathtools}
\usepackage{amsthm}
\usepackage[capitalize]{cleveref}
\usepackage{algpseudocode}
\usepackage{graphicx}
\usepackage{enumitem}
\usepackage{xcolor}
\usepackage{color,soul}
\usepackage{tikz} 
\usepackage{mathabx}
\usepackage{physics}
\usepackage{bbm}
\usepackage[ruled,vlined]{algorithm2e}
\usepackage{subfigure}
\usepackage{dsfont}
\ecaisubmission

\newtheorem{assumption}{Assumption}
\newtheorem{remark}{Remark}
\newtheorem{lemma}{Lemma}
\newtheorem{theo}{Theorem}

\begin{document}

\begin{frontmatter}

\title{Cooperative Thresholded Lasso for Sparse Linear Bandit}

\author[A]{\fnms{Haniye}~\snm{Barghi}}
\author[A]{\fnms{Xiaotong}~\snm{Cheng}}
\author[A]{\fnms{Setareh}~\snm{Maghsudi}}

\address[A]{Eberhard Karls University of Tübingen, Tübingen, Germany \\
\{haniyeh.barghi, xiaotong.cheng, setareh.maghsudi\}@uni-tuebingen.de}

\begin{abstract}
We present a novel approach to address the multi-agent sparse contextual linear bandit problem, in which the feature vectors have a high dimension $d$ whereas the reward function depends on only a limited set of features - precisely $s_0 \ll d$. Furthermore, the learning follows under information-sharing constraints. The proposed method employs Lasso regression for dimension reduction, allowing each agent to independently estimate an approximate set of main dimensions and share that information with others depending on the network's structure. The information is then aggregated through a specific process and shared with all agents. Each agent then resolves the problem with ridge regression focusing solely on the extracted dimensions. We represent algorithms for both a star-shaped network and a peer-to-peer network. The approaches effectively reduce communication costs while ensuring minimal cumulative regret per agent. Theoretically, we show that our proposed methods have a regret bound of order $\mathcal{O}(s_0 \log d + s_0 \sqrt{T})$ with high probability, where $T$ is the time horizon. To our best knowledge, it is the first algorithm that tackles row-wise distributed data in sparse linear bandits, achieving comparable performance compared to the state-of-the-art single and multi-agent methods. Besides, it is widely applicable to high-dimensional multi-agent problems where efficient feature extraction is critical for minimizing regret. To validate the effectiveness of our approach, we present experimental results on both synthetic and real-world datasets.
\end{abstract}

 \end{frontmatter}

\section{Introduction}
Cooperative multi-agent bandit is a suitable framework to tackle complex decision-making problems across a broad spectrum of applications such as Ad-Hoc networks \cite{7498076}, personalized recommendation systems \cite{210608902}, traffic management \cite{b2bcfa5d95c349b4b79272d770ab1a7e}, and the like. In such a framework, the challenge is to enable each agent to learn from its own experiences while considering the actions and rewards of other agents in the system. Given the limitations imposed by the environment, it is crucial to simultaneously keep communication between agents to a minimum during the learning process. To simplify the bandit problems with a large set of arms, it is common to assume a specific model for the payoff functions \cite{slivkins2011contextual}, e.g., the linear structure between actions and rewards \cite{li2019improved}. 

The state-of-the-art research about multi-agent linear bandit problems seldom considers the high dimensional action space \cite{10.5555/3524938.3525195, dubey2020differentially}. The dimension of the action space accounts for a dominant part in both regret bound \cite{kim2019doubly} and communication cost \cite{9786669, 1604.07706}. Real-world settings often entail noisy components comprising web or mobile-based contexts \cite{kim2019doubly,cella2021multi}, while most relevant features are small and yield a sparse model parameter. The main challenge in sparse linear bandits is learning the sparse structure of the reward function, as only a small subset of features are relevant for prediction, whereas others are irrelevant or noisy. By relying on prior knowledge or presumptions about the sparsity structure, the sparse linear bandit framework offers a potent mathematical model to address this challenge \cite{JMLR:v19:17-025, pmlr-v22-abbasi-yadkori12, oh2021sparsity}.

We propose a novel multi-agent linear bandit algorithm that handles high-dimensional action space when only a minor subset of dimensions is related to the reward. Despite its versatility, a multi-agent version of sparse linear bandits remains unexplored. We develop a collaborative information-sharing mechanism with low communication cost that assists the agent in fast and accurate estimation of the support set of the sparse parameter. To the best of our knowledge, only \cite{9786669} tackles decentralized sparse bandits; Nevertheless, compared to our model, it includes several limiting assumptions. Specifically, our cooperative framework integrates information sharing among agents into the high-dimensional linear bandit algorithm. Besides, it does not require any prior knowledge regarding the sparse structure. Our main contributions are summarized as follows.
\begin{itemize}
    \item \textbf{The CTL Algorithm:} 
    We propose an innovative solution, namely, the Cooperative Thresholded Lasso Linear bandit (CTL) algorithm, for the multi-agent sparse linear bandit problem. Our proposal leverages the combination of ridge estimation for arm selection and parameter estimation and thresholded Lasso bandit for dimension reduction. We consider two variants of in-network communication: i) centralized framework, where a central server node aggregates the information from all agents and then distributes the results to them, and ii) decentralized peer-to-peer framework, where agents communicate directly with each other without coordination of the central server node. To reduce the communication burden, we propose a communication framework that reduces the total communication rounds to $\mathcal{O}(\log T)$. Our algorithm is simple and easily generalizable, meaning that it can accommodate other dimension reduction techniques or different communication network topologies using little adaptations. That remarkable robustness and flexibility make CTL an attractive solution for multi-agent sparse linear bandit problems.
    
    \item \textbf{Performance Evaluation:} We establish that the high probability group regret bound of our proposed CTL algorithm is $\mathcal{O}(s_0 \log d + s_0 \sqrt{T})$, where $T$, $s_0$ and $d$ refer to the number of time steps, non-zero elements in the feature vector and the dimension of the feature vector, respectively. Besides, we prove that the total communication cost is $\mathcal{O}(s_0 \log T)$. These bounds show that CTL is a practical solution for multi-agent sparse linear bandit problems that balances the trade-off between communication and computation cost while retaining low cumulative regret. 
    
    \item \textbf{Numerical Experiment:} We demonstrate the efficacy of the CTL algorithm through extensive numerical experiments. We compare our proposed algorithm with a series of state-of-the-art sparse linear bandit algorithms, including the Thresholded Lasso \cite{ariu2022thresholded}, Sparsity-Agnostic Lasso \cite{oh2021sparsity}, and Doubly-Robust Lasso \cite{kim2019doubly}. Experiments on synthetic- and real-world datasets show the superior performance of our proposal. The results highlight the advantages of utilizing a multi-agent framework compared to a single agent with the same number of observations. Besides, we compare our method with a multi-agent low-dimensional algorithm \cite{9786669}, effectively showing the superiority of our approach. The CTL algorithm outperforms the referenced method not only in terms of cumulative regret but also by imposing significantly fewer simplification assumptions.
\end{itemize}
\subsection{Related Works}
Our work is closely related to the research on multi-agent linear bandits and sparse linear bandits. In this section, we  review the state-of-the-art research in both directions and then highlight connections between them. 

Multi-agent bandit problem has gained great attention in the past few years \cite{10.1145/3571306.3571432, Amani_Thrampoulidis_2021, pmlr-v119-dubey20a}. The proposed strategies for multi-agent problems are dividable into two main categories. Most related works assume that agents continually select from a limited subset of arms and exchange their beliefs about the best arm in their playing set \cite{ijcai2017p24, pmlr-v108-chawla20a, 007.03812}. Reference \cite{ijcai2017p24} proposes a teamwork model in which the agents decide whether to pull the arms of a bandit or broadcast their obtained rewards over several epochs, aiming to maximize the total rewards. The model captures a three-way tradeoff between exploration, exploitation, and communication. They also show that the proposed decentralized algorithm with a Value-of-Information communication strategy converges rapidly to the performance of a centralized method. However, in our research and some others, all agents face the same environment; That is, they share the entire set of arms among agents. Thus, the algorithm must consider all arms at each time step. Our research is tightly related to the literature in multi-agent linear bandits such as \cite{1904.06309, 1604.07706}. Wang et al. \cite{1904.06309} present a communication-efficient algorithm under the coordination of a central server, allowing every agent to have immediate access to the complete full network information. Compared to \cite{1904.06309}, our proposed algorithm has both a centralized and decentralized structure; Hence it has a wide range of applications. Reference \cite{1604.07706} studies distributed linear bandits in peer-to-peer networks, where each agent can only send information to one randomly chosen agent per round. We consider centralized and decentralized communication networks and allow for less frequent communications. Besides, to our best knowledge, previous research rarely considers the sparse parameter. 

Another line of research related to ours is the sparse linear bandit problem. Some papers in that direction assume the availability of side information about the sparse model parameters \cite{pmlr-v22-abbasi-yadkori12, gilton2017sparse, kim2019doubly}. For example, to tackle the sparse linear stochastic bandit problem, \cite{pmlr-v22-abbasi-yadkori12} introduces a technique, namely, online-to-confidence-set conversion, to construct high-probability confidence sets for linear prediction with correlated inputs. However, it requires the sparsity level of the model, i.e., the size of the support set. Reference \cite{gilton2017sparse} leverages ideas from linear Thompson sampling and relevance vector machines, resulting in a scalable approach that adapts to the unknown sparse support. That paper also assumes prior knowledge of a slightly larger set of support for the model parameter. Recently, studies on sparse linear bandits overcome that limitation \cite{oh2021sparsity, ariu2022thresholded}, which do not require any prior information about the sparse parameter of the model. Moreover, thresholding has become a natural and efficient way to feature selection in online and offline learning \cite{ariu2022thresholded, Zhou2010ThresholdedLF, 9079458}, which achieves excellent performance in sparse linear bandit. Thus, establishing a dependable interval for the threshold value is necessary and crucial for the algorithm to operate effectively. The closest work to our setting is \cite{ariu2022thresholded}, which uses the Lasso framework with thresholding to maintain and update the estimates about the support set of the model parameter.

The paper structure is as follows: Firstly, in Section \ref{sec:problem}, we provide a formal statement of the problem. Then, we discuss the algorithm for centralized and peer-to-peer settings in Section \ref{sec:algorithm}. In Section \ref{sec:rgrt_bnd}, we establish a regret bound for our proposed algorithm. We evaluate our proposal numerical using synthetic- and real datasets in Section \ref{sec:expr}.

\section{Problem Formulation} 
\label{sec:problem}
\subsection{Model and Notation}
We consider a multi-agent high-dimensional linear bandit problem with $N$ agents. Let $T$ become the problem horizon, i.e., the number of rounds to be played. At each time step $t \in [T]$, each agent $i \in [N]$ receives a set of $K$ context vectors $\mathcal{A}_t^i \subset \mathbb{R}^{K \times d}$ sampled from one unknown distribution. Each agent $i$ selects an action $A_t^i \in \mathcal{A}_t^i$ based on the previous observations in round $t$ and obtains the reward $y_t^i$,
\begin{gather}
    y_t^i:= \langle A_t^i , \theta^* \rangle + \omega_t^i,
\end{gather}
where $\theta^* \in \mathbb{R}^d$ is an unidentified sparse parameter, and $\omega_t^i$ is sub-Gaussian noise with a zero mean. Parameter $\theta^*$ and $A_t^i$ are both high-dimensional $d \gg 1$, while parameter $\theta^*$ is sparse, which means the number of non-zero elements $s_0 = \norm{\theta^*}_0 \ll d$. In other words, $\theta^*$ is $s_0$-sparse and $s_0$ is a constant but unknown integer. Furthermore, if $\mathcal{F}_t^i$ is the $\sigma$-algebra generated by random variables $(\mathcal{A}_1^i, A_1^i, y_1^i, \ldots, A_{t-1}^i, y_{t-1}^i, \mathcal{A}_t^i)$, $A_t^i$ is $\mathcal{F}_t^i$-measurable. The noise term $\omega_t^i$ is independent across agents given $\mathcal{F}_t^i$ and $A_t^i$. Moreover, we have the sub-Gaussian property such that $\mathbb{E}[e^{\alpha \omega_t^i}] \leq e^{\alpha^2 \sigma^2/2}, \forall \alpha \in \mathbb{R}$, where $\sigma$ is a positive constant. This inequality implies that the moment-generating function of $\omega_t^i$ exists and is bounded, which is a desirable property in many statistical and mathematical models.

At time step $t$, the instantaneous expected regret of each agent $i \in [N]$ yields 
\begin{align*}
    r_t^i := \mathbb{E} [ \max_{A\in \mathcal{A}_t^i} \langle A - A_t^i, \theta^* \rangle].
\end{align*}
The cumulative regret for any agent $i$ is $R_i(T) := \sum_{t=1}^T  r_t^i$. The objective of each agent is to minimize its overall cumulative regret as an individual.

\paragraph{Notation} 
The $\ell_0$-norm of a vector $x \in \mathbb{R}^d$ is $\norm{x}_0= \sum_{j = 1}^d  \mathbbm{1} \left\{ x_j \neq 0\right\}$. The set $S(x) := \{j \in [d] = \{1, 2, \ldots, d\} : x_j \neq 0\}$ stands for the support of a vector $x$. For each agent $i$, the empirical Gram matrix that the arms produced under a certain algorithm is represented by $\hat{\Sigma}_{t,i} = \frac{1}{t}\sum_{s=1}^t A_{s,i} A_{s,i}^\top$. For any $B \subset [d]$, we define $x_B \coloneqq (x_{1, B}, \ldots, x_{d, B})^\top $ where for all $j \in [d]$, $x_{j, B} \coloneqq x_j \mathbbm{1} \{ j \in B\}$. Additionally, we define $x_{\min}$ as $|x_j|$'s minimal value on its support: $x_{\min} \coloneqq \min_{j \in S(x)} |x_j|$. The weighted norm-2 of vector $x \in \mathbb{R}^d$ is defined as $\norm{x}_A := \sqrt{x^\top A x}$, where $A\in \mathbb{R}^{d\times d}$ is a positive definite matrix. We define the minimum eigenvalue of a matrix $A$ as $\lambda_{\min}(A)$.

\subsection{Assumptions}
Below, we outline our assumptions that mostly stem from \cite{oh2021sparsity, ariu2022thresholded}, and compare them to those in the related literature. 
\begin{assumption}[Context vector and parameter constraints] \label{asm:par_cons}
For the feature vector $\theta^*$, we assume that $\norm{\theta^*}_1 \le s_1$ for some unknown constant $s_1$ and $\norm{\theta^*}_2 \le s_2$, where $s_2$ is a positive constant. Besides, we assume that the context vector's $\ell_{\infty}$-norm is bounded: for all $t$, $i \in [N]$ and for all $A \in \mathcal{A}_t^i$, $\norm{A}_{\infty} \le s_A$, where $s_A > 0$ is a constant.
\end{assumption}
Bounded norms of model parameter and feature vectors are common assumptions in high dimensional linear models \cite{kim2019doubly, 10.5555/3305890.3305895}.
\begin{assumption}[Compatibility condition]
\label{asm:comp_cond}
We specify the compatibility constant $\phi(M, S_0)$ as
\begin{align*}
\phi^2(M, S_0) \!\coloneqq \!\min_{x:  \|x_{S_0}\|_1 \neq 0} \left\{\! \frac{s_0 x^\top M x}{\|x_{S_0}\|_1^2} \!:\! \|x_{S_0^c}\|_1 \!\le\! 3 \|x_{S_0}\|_1 \!\right\}\!
\end{align*}
for a matrix $M \in \mathbb{R}^{d \times d}$ and a set $S_0 \subset [d]$. We assume that for the Gram matrix of the action set $\Sigma \coloneqq \frac{1}{K} \sum_{k=1}^K \mathbb{E}_{\mathcal{A}\sim p_A} \left[A_{k} A_{k}^\top\right]$ satisfies $\phi^2(\Sigma, S(\theta^*)) \ge \phi^2_0$, where $\phi_0$ is some positive constant. 
\end{assumption}
In the high dimensional statistics literature, the compatibility condition appeared for the first time in \cite{buhlmann2011statistics}. It is similar to the standard Gram matrix positive-definiteness for the ordinary least square estimator for linear models, but less constricting. The compatibility condition ensues that the parameter's truly active components are not strongly correlated. According to many pertinent studies, Assumption \ref{asm:comp_cond} is essential for the consistency of the Lasso estimation.
\begin{assumption}[Relaxed symmetry \cite{oh2021sparsity}]\label{asm:relax_sym}
For the distribution $p_A$ of $\mathcal{A}$, there exists a constant $\nu\ge1$ such that for all $\vec{A} \in \mathbb{R}^{K \times d}$ with $p_A(\vec{A})>0$, $\frac{p_A(\vec{A})}{p_A(- \vec{A})} \le \nu$.
\end{assumption}
Assumption~\ref{asm:relax_sym} stems from \cite{oh2021sparsity}. According to this assumption, the joint distribution $p_A$ may exhibit skewness, but this skewness is subject to some constraints. It is known that a broad class of continuous and discrete distributions, such as Gaussian distributions, multi-dimensional uniform distributions, and Rademacher distributions, satisfy the property of relaxed symmetry. This property ensures that the distribution remains symmetric even in the presence of small deviations from the perfect symmetry, allowing for some degree of skewness while still maintaining overall balance.
\begin{assumption}[Balanced covariance \cite{oh2021sparsity}]
\label{asm:balanced_cov}
For any permutation $\gamma$ of $[K]$, for any integer $k \in \{2, \ldots,K-1\}$ and a fixed $\theta^*$, there exists a constant $C_{\textnormal{b}}> 1$ such that
\begin{align*}
&C_{\textnormal{b}}  \mathbb{E}_{\mathcal{A} \sim p_A} \big[ ( A_{\gamma(1)} A^\top_{\gamma(1)} + A_{\gamma(K)} A^\top_{\gamma(K)})\\
& \qquad \qquad \quad \cdot \mathds{1} \{ \langle A_{\gamma(1)},  \theta^* \rangle < \ldots < \langle A_{\gamma(K)}, \theta^* \rangle \} \big]\\
&\succeq \mathbb{E}_{\mathcal{A} \sim p_A}\left[ A_{\gamma(k)} A^\top_{\gamma(k)}  \mathds{1} \{ \langle A_{\gamma(1)}, \theta^* \rangle < \ldots < \langle A_{\gamma(K)}, \theta^* \rangle \} \right].
\end{align*}
\end{assumption}
We adapted Assumption~\ref{asm:balanced_cov} from \cite{oh2021sparsity}. The statement is valid for a variety of distributions, such as multivariate Gaussian distribution and uniform distribution on the sphere. It still applies when contexts are independent of one another with any arbitrary distributions \cite{oh2021sparsity}.
\begin{assumption}[Sparse positive definiteness]
\label{asm:pos-def}
For each $B \subset [d]$, define $\Sigma_B = \frac{1}{K}\sum_{k=1}^K \mathbb{E}_{\mathcal{A}\sim p_A}[A_k(B)A_k(B)^\top]$, where $A_k(B)$ is a $|B|$-dimensional vector, which is extracted from the elements of $A_k$ with indices in $B$. There exists a positive constant $\alpha > 0$ such that $\forall B \subset [d]$,
\begin{align*}
    |B| \leq s_0 &+ (4\nu C_b s_0)/\phi_0^2 \quad \textup{and} \quad S(\theta^*) \subset B \notag \\ 
    & \Rightarrow \min_{v \in \mathbb{R}^{|B|}:\norm{v}_2 = 1} v^\top\Sigma_B v \geq \alpha. 
\end{align*}
\end{assumption}
The parameters $\phi_0$, $\nu$, and $C_b$ match those of Assumption~\ref{asm:comp_cond}, \ref{asm:relax_sym}, and \ref{asm:balanced_cov}. According to Assumption~\ref{asm:pos-def}, the context distribution around the support of $\theta^*$ is sufficiently diverse. In low dimensional linear bandit literature, Assumption~\ref{asm:pos-def} is commonly used (e.g., \cite{lattimore2020bandit, Degenne2020GamificationOP, NEURIPS2020_7212a656}).
\section{Algorithm} 
\label{sec:algorithm}
In this section, we present the Cooperative Thresholded Lasso bandit algorithm (CTL), which adapts the concept of thresholding in \cite{ariu2022thresholded} and the LinUCB algorithm \cite{abbasi2011improved}. In this method, each agent selects an action based on an estimate of the feature vector $\theta^*$ at each time step $t$. The estimation follows from two main working components, ridge regression, and the thresholded Lasso. Instead of computing the decision-making policy in high-dimensional space $d$, with the help of thresholded Lasso, the agents decide in a space with a ``reduced'' number of dimensions, which reduces the computational cost significantly.

\begin{algorithm}
\SetAlgoLined
\SetAlgoHangIndent{1.5em}
\SetArgSty{textnormal}
\caption{Centralized Cooperative Thresholded Lasso Bandit  Algorithm (CCTL)}
\label{alg:CCTL}
    \textbf{initialisation:} $\lambda_0$, $\xi$, $\hat{S}_1 = \{1, \ldots, d\}$, and $\forall i \in [N]: M_1^i = I_{d \times d}$, $b_1^i = 0_{1 \times d}$ \\
    \For{$t = 1, 2, \ldots, T$}{
		\For{agent $i \in [N]$}{
            $\hat{\theta}^i_{t} \leftarrow (M_{t}^i)^{-1}b_{t}^i$ \\
			Observe context vectors of all arms $\mathcal{A}_t^i \in \mathbb{R}^{K \times d}$ \\
			$\tilde{\mathcal{A}}_t^i \leftarrow $ remove dimensions $[d] \setminus \hat{S}_t$ from $\mathcal{A}_t^i$ \\
			Select $k' = \arg \max_{k \in [K]} \langle \tilde{\mathcal{A}}_{t,k}^i, \hat{\theta}^i_t \rangle$, observe reward $y_t^i$, and $A_t^i = \tilde{A}_{t,k'}^i$ \\
			Add $\mathcal{A}_{t,k}^i$ to $A_i$ and $y_t^i$ to $Y_i$ \\
			Update weights $M_{t+1}^i = M_{t}^i + A_t^i (A_t^i)^\top$ and $b_{t+1}^i = b_{t}^i + y_t^i A_{t}^i$
        }
		\If{$\log_\xi(t) \in \mathbb{N}$}{
			$\lambda_t \leftarrow \lambda_0 \sqrt{\frac{2\log t \log d}{t}}$ \\
			$\mathcal{T}_t \leftarrow N \times \lambda_t$ \\
			\For{agent $i \in [N]$}{
				$\hat{\theta}^i_t \leftarrow \arg \min_{\theta}\{\frac{1}{t}\norm{Y_i - \langle A_i,\theta \rangle}_2^2 + \lambda_t \norm{\theta}_1\}$ \\
				$\hat{S}_{t+1}^i \leftarrow \{j \in [d]: |(\hat{\theta}_t^i)_j| > \mathcal{T}_t\}$
			}
			\textbf{server:} $\hat{S}_{t+1} \leftarrow  \underset{i=1}{\overset{N}{\bigcup}} \hat{S}_{t+1}^i$ \\
			\textbf{each agent $i \in [N]$}: Update $M_{t+1}^i$ and $b_{t+1}^i$ according to $\hat{S}_{t+1}$ \\}
		\Else{
			$\hat{S}_{t+1} \leftarrow\hat{S}_t$}
	}
\end{algorithm}

In the federated setting, where agents have different actions and estimations, the communication protocol design is critical. In this section, we first consider a centralized communication framework where there exists a centralized server node that coordinates the communication among agents. Under the centralized communication protocol, each agent periodically communicates with the centralized server and synchronizes itself with other agents. We then extend the framework to a decentralized peer-to-peer network setting. Theoretical guarantees are provided for the performance of both structures.

\subsection{Centralized Framework with a Server Node}

Algorithm \ref{alg:CCTL} summarizes the centralized version of CTL (CCTL), which operates as follows. Initially, each agent $i$ assigns $M_1^i = I_{d \times d}$ and $b_1^i = 0_{1 \times d}$ for use in ridge regression. Additionally, $\hat{S}_t$ provides an estimate of the support set of $\theta^*$ and is initialized with $\hat{S}_1 = \{1,\ldots,d\}$, including all dimensions. At each step $t$, every agent chooses an action optimistically based on the estimated $\hat{\theta}_t^i$, while only considering the dimensions provided in $\hat{S}_t$. After receiving the reward, each agent updates its estimate based on ridge regression. During the synchronization step, when $\log_\xi t \in \mathbb{N}$ and $a > 1$, agents obtain an estimate via Lasso, which is used to estimate the support of $\theta^*$ with appropriate thresholding computation. We select the regularizer based on the setting in \cite{ariu2022thresholded}. Unlike \cite{ariu2022thresholded}, here we only perform one threshold procedure. To save communication costs, agents only share their estimate of $\theta^*$'s support. After synchronization, the server node obtains the final estimate of $\hat{S}_t$ by taking the union of the support sets of the shared sets.

\begin{algorithm}
\SetAlgoLined
\SetAlgoHangIndent{1.5em}
\SetArgSty{textnormal}
\caption{Decentralized Peer-to-Peer Cooperative Thresholded Lasso Linear Bandit Algorithm (DCTL)}\label{alg:DCTL}
    \textbf{initialisation:} $\lambda_0$, $\xi$, $\hat{S}_1^i = \{1, \ldots, d\}$, and $\forall i \in [N]: M_1^i = I_{d \times d}$, $b_1^i = 0_{1 \times d}$ \\
    \For{$t = 1, 2, \ldots, T$}{
		\For{agent $i \in [N]$}{
            $\hat{\theta}^i_{t} \leftarrow (M_{t}^i)^{-1}b_{t}^i$ \\
			Observe context vectors of all arms $\mathcal{A}_t^i \in \mathbb{R}^{K \times d}$ \\
			$\tilde{\mathcal{A}}_t^i \leftarrow $ remove dimensions $[d] \setminus \hat{S}_t^i$ from $\mathcal{A}_t^i$ \\
			Select $k' = \arg \max_{k \in [K]} \langle \tilde{\mathcal{A}}_{t,k}^i, \hat{\theta}^i_t \rangle$, observe reward $y_t^i$, and $A_t^i = \tilde{A}_{t,k'}^i$ \\
			Add $\mathcal{A}_{t,k}^i$ to $A_i$ and $y_t^i$ to $Y_i$ \\
			Update weights $M_{t+1}^i = M_{t}^i + A_t^i (A_t^i)^\top$ and $b_{t+1}^i = b_{t}^i + y_t^i A_{t}^i$
        }
		\If{$\log_\xi(t) \in \mathbb{N}$}{
			$\lambda_t \leftarrow \lambda_0 \sqrt{\frac{2\log t \log d}{t}}$ \\
			$\mathcal{T}_t \leftarrow 2 \times \lambda_t$ \\
			\For{agent $i \in [N]$}{
				$\hat{\theta}^i_t \leftarrow \arg \min_{\theta}\{\frac{1}{t}\norm{Y_i - \langle A_i,\theta \rangle}_2^2 + \lambda_t \norm{\theta}_1\}$ \\
				$\hat{S}_{t+1}^i \leftarrow \{j \in [d]: |(\hat{\theta}_t^i)_j| > \mathcal{T}_t\}$ \\
                Select agent $j \in \mathcal{N}_i$ to communicate and obtain its estimate $\tilde{S}_{t+1}^j$ \\
                $\hat{S}_{t+1}^i \leftarrow  \tilde{S}_{t+1}^i \bigcup \tilde{S}_{t+1}^j$ \\
                Update $M_{t+1}^i$ and $b_{t+1}^i$ according to $\hat{S}_{t+1}^i$
			}
        }
		\Else{
			$\hat{S}_{t+1} \leftarrow\hat{S}_t^i$}
	}
\end{algorithm}

\begin{remark}
Similar to \cite{McMahan2016CommunicationEfficientLO}, the algorithm above is generalizable by selecting a random subset of agents in each synchronization step. That approach allows for more flexibility in network coverage, particularly in scenarios where not all agents are consistently online in the system. Additionally, it enables the management of large-scale systems in which many agents are involved, and limited communication capacity is a potential bottleneck. By carefully selecting only a subset of agents to participate in each synchronization round, the algorithm can effectively balance communication demands with the computational and operational capabilities of the system while still maintaining a high degree of accuracy in the estimation of $\hat{S}_t$.
\end{remark}

\subsection{Decentralized Peer-to-Peer Framework}
In this scenario, each agent communicates directly with its neighbors via a decentralized peer-to-peer protocol. The communication network is modeled by an undirected network $G = (N,E)$, where $e_{i,j} \in E$ if agent $i$ and $j$ can communicate directly, or in other words, $i$ and $j$ are neighbors. Define $\mathcal{N}_i$ as the neighbors of agent $i$. At each synchronization step, the algorithm proceeds as follows: Once each agent obtains the estimation of $\theta^*$'s support, it randomly selects a neighbor and receives the corresponding support's estimation of that selected neighbour. This additional information is then integrated into the agent's own support estimation through a union operation, enabling the agent to enhance the recall and robustness of its estimate. Here, recall refers to the probability of the main dimensions appearing in the support estimation. All other steps in the algorithm remain similar to those described in the centralized version. Algorithm~\ref{alg:DCTL} summarizes the Decentralized peer-to-peer CTL algorithm (DCTL).

\section{Performance Analysis} 
\label{sec:rgrt_bnd}
\subsection{Centralized Framework}
\begin{theo}
\label{theo:CenRegret}
		Consider a system consisting of $N$ agents connected by a server node. Every agent uses Algorithm \ref{alg:CCTL} to select arms in each time step. Under Assumption \ref{asm:par_cons}-Assumption \ref{asm:pos-def}, we can establish the existence of a positive constant $c$ such that $\lambda_0 = 4\sqrt{c}\sigma s_A$. Then, for all $d \ge \exp(4/c)$ and $T \ge 2$, with probability at least $(1-\delta)$, the following inequality holds:
		\begin{align*}
            R_i(T) &\leq 2s_As_1\tau + \frac{8K\sqrt{\xi}}{\sqrt{\xi} - 1} (\sqrt{\xi T} - 1) \\
            &\quad \quad \quad \sqrt{
            \begin{aligned}
                \sigma^2 & C_a^2 (s_0 + \frac{16 s_0 \nu C_b}{\phi_0})^2 \log^2 T + \\ & (s_2^2 - 2 \sigma^2 \log \delta) C_a (s_0 + \frac{16 s_0 \nu C_b}{\phi_0})\log T
            \end{aligned}
            }\\
    	    &+ 2 K s_A s_1 \left( \frac{1-T^{1-2N}}{2N-1} + \frac{4}{N C_0^2} + (s_0 + \frac{16\nu C_b s_0}{\phi_0^2})^2 \right.\\ 
            & \qquad \qquad \qquad \left. \frac{40 s_A \nu C_b}{\alpha} \right),
        \end{align*}
\end{theo}
where $\tau = \max \big\{\frac{2\log(2d^2)}{C_0^2}, \exp(2\log \xi + \frac{2}{c})\big\}.$

\begin{proof}[Proof sketch] We outline the proof of Theorem~\ref{theo:CenRegret} as follows.
\begin{itemize}
    \item \textbf{Performance Analysis of Estimated Support Set:} Given that Algorithm \ref{alg:CCTL} iteratively reduces the dimension, the initial step in evaluating the regret bound for our proposed method entails assessing the estimated support set following each synchronization round. To this end, we present the following Lemma to give a tight lower bound for the probability of the existence of $S(\theta^*)$ and the extent of false positive features in this estimation $(\hat{S}_t)$. We prove Lemma \ref{lem:cen-supp} in Appendix \ref{app:prf-supp}.
    \begin{lemma} 
    \label{lem:cen-supp}
        (Centralized Framework) Assume that, for each agent $i \in [N]$, assumptions \ref{asm:par_cons}, \ref{asm:comp_cond}, \ref{asm:relax_sym}, and \ref{asm:balanced_cov} hold. Then for all $t \ge \frac{2\log(2d^2)}{C_0^2}$ with $C_0 := \min\{\frac{1}{2}, \frac{\phi_0^2}{512s_0 s_A^2 \nu C_b}\}$ the event $\mathcal{E}_t = \{|\hat{S}_t \setminus S(\theta^*)| \le \frac{16 s_0 \nu C_b}{\phi_0^2} \text{ and } S(\theta^*) \subset \hat{S}_t \}$ holds true with probability at least
    	\begin{align*}
        	1 - \left(2\exp(-\frac{t'\lambda_{t'}^2}{32\sigma^2s_A^2}+\log d)\right)^N - \exp(-\frac{Nt'C_0^2}{2}),
            \end{align*}
    	where $t' = \xi^{\lfloor \log_\xi t \rfloor}$.
    \end{lemma}
    Lemma \ref{lem:cen-supp} represents an extension of the support recovery outcome of the Thresholded Lasso Bandit (as stated in \cite{ariu2022thresholded}) to the circumstance of multiple agents exchanging information among each other. The reliance on $s_0$ instead of $d$ is similar to that of the offline result (as Theorem 3.1 of \cite{Zhou2010ThresholdedLF}) and the bandit setting illustrated in Lemma 5.4 of \cite{ariu2022thresholded}. Our thresholding approach, combined with the allowance of agents to share their estimated sets, facilitates a more precise dimension reduction through the learning process, effectively removing the reliance on $d$ for estimation error when $t$ exceeds $2\log(2d^2)/C_0^2$. This, in turn, leads to improved regret bounds as compared to those established in existing literature, such as \cite{oh2021sparsity} or \cite{ariu2022thresholded}.
        
    \item \textbf{Minimal Eigenvalue of the Empirical Gram Matrix:}
    We introduce the notion of $\hat{\Sigma}_{t,i}$ as the empirical Gram matrix on the estimated support of agent $i$, up to time step $t$. This matrix is a fundamental tool to capture the pairwise relationships between estimated survival probabilities at different time points. The desirable property of positive definiteness of $\hat{\Sigma}_{t,i}$ ensures that it is not only invertible but also allows for the utilization of powerful mathematical tools for statistical inference. Our proposed lemma aims to establish the positive definiteness of $\hat{\Sigma}_{t,i}$, even when the underlying data generating process is not i.i.d. Notably, this lemma shares similarities with Lemma 5.6 presented in \cite{ariu2022thresholded}.
        \begin{lemma} 
        \label{lem:egnval}
        Under Assumptions \ref{asm:par_cons} and \ref{asm:pos-def}, for any agent $i\in [N]$ and for all $t\in [T]$, we have: 
        \begin{gather*}
            \mathbb{P}(\lambda_{min}(\hat{\Sigma}_{t,i}) \ge \frac{\alpha}{4 \nu C_b}|\mathcal{E}_t) \ge \\
            1 - \exp(\log(s_0 + \frac{16 s_0 \nu C_b}{\phi_0^2}) - \frac{t' \alpha}{20 s_A \nu C_b (s_0 + \frac{16 s_0 \nu C_b}{\phi_0^2})}),
        \end{gather*}
        where $t' = \xi^{\lfloor \log_\xi t \rfloor}$.
        \end{lemma}
        The proof of this Lemma is similar to that of Lemma 5.6 in \cite{ariu2022thresholded}, albeit with a minor change. Mainly, in the utilization of Lemma F.10 \cite{ariu2022thresholded}, we must modify the upper bound for the size of estimated support set $\hat{S}_t$ to $s_0 + (16 s_0 \nu C_b)/\phi_0^2$, while retaining all other steps unchanged.
        
    \item \textbf{Instantaneous Regret Upper Bound:}
        Below, we state a lemma that serves to bound the instantaneous regret for each agent $i \in [N]$. We prove this lemma in Appendix \ref{app:prf-inst-rgt} based on \cite{abbasi2011improved}.
        \begin{lemma}
        \label{lem:inst-rgt}
		For any $t \in [T]$ and each agent $i\in [N]$, with probability at least $1-\delta$ the instantaneous regret $r_t^i = \mathbb{E}[\max_{A\in \mathcal{A}_t^i}\langle A - A_t^i,\theta^* \rangle]$ is upper bounded as
		\begin{gather*}
			r_t^i \le \sum_{k=1}^K \mathbb{E} \left[(\norm{A_{t,k}^i}_{(M_t^i)^{-1}}+\norm{A_t^i}_{(M_t^i)^{-1}})(\sigma \sqrt{\log(\frac{\det(M_t^i)}{\delta^2})} \right. \\ 
            \left. + \norm{\theta}^*_2) |\mathcal{A}_t^i \in \mathcal{R}_k^i, \mathcal{E}_t, \mathcal{G}_{t,i}^{\frac{\alpha}{4 \nu C_b}}\right] \\
            + 2 K s_A s_1 \big(\mathbb{P}((\mathcal{E}_t)^c) + \mathbb{P}((\mathcal{G}_{t,i}^{\frac{\alpha}{4 \nu C_b}})^c | \mathcal{E}_t)\big),
		\end{gather*}
		where $\mathcal{R}_k^i := \{\mathcal{A}_t^i \in \mathbb{R}^{K \times d}: k \in \arg \max_{k'}\langle A_{t,k'}^i, \theta^* \rangle\}$ and $\mathcal{G}_{t,i}^{\lambda} := \{\lambda_{\min}(\hat{\Sigma}_{\hat{S}_t}^i) \geq \lambda\}$.
        \end{lemma}
\end{itemize}
With the aforementioned lemmas, we can prove Theorem \ref{theo:CenRegret}, as provided in Appendix \ref{app:prv-CenRegret}.
\end{proof}
\subsection{Decentralized Peer-to-Peer Framework}
\begin{theo}
\label{theo:deRegret}
    Consider a network of $N$ agents connected via a fix connected graph. In each time step, the system chooses arms using the algorithm \ref{alg:DCTL}. There is a positive constant $c$ such that $\lambda_0 = 4\sqrt{c}\sigma s_A$ under the necessary conditions of \ref{asm:par_cons}-\ref{asm:pos-def}. We hereby declare that the following inequality holds true with a probability of at least $1-\delta$ for any $d \ge \exp(4/c)$ and for all $T \ge 2$
		\begin{align*}
            R_i(T) &\leq 2s_As_1\tau + \frac{8K\sqrt{\xi}}{\sqrt{\xi} - 1} (\sqrt{\xi T} - 1) \\
            &\quad \quad \quad \sqrt{
            \begin{aligned}
                \sigma^2 & C_a^2 (s_0 + \frac{16 s_0 \nu C_b}{\phi_0})^2 \log^2 T + \\ & (s_2^2 - 2 \sigma^2 \log \delta) C_a (s_0 + \frac{16 s_0 \nu C_b}{\phi_0})\log T
            \end{aligned}
            }\\
    	    &+ 2 K s_A s_1 \left( \frac{1-T^{1-2N}}{2N-1} + \frac{4}{N C_0^2} + (s_0 + \frac{16\nu C_b s_0}{\phi_0^2})^2 \right. \\
            & \qquad \qquad \qquad \left. \frac{40 s_A \nu C_b}{\alpha}\right),
        \end{align*}
\end{theo}
where $\tau = \max \big\{\frac{2\log(2d^2)}{C_0^2}, \exp(2\log \xi + \frac{2}{c})\big\}.$
\begin{remark}
The proof of Theorem \ref{theo:deRegret} is almost identical to that of the centralized version with a notable difference. Specifically, it pertains to the communication process, whereby each agent interacts exclusively with an agent at every step. Consequently, it behooves us to assign $N = 2$ in \eqref{second} and proceed with the remaining steps in a similar manner.
\end{remark}
\begin{figure*}[!t]
    \centering
    \includegraphics[width=0.4\linewidth]{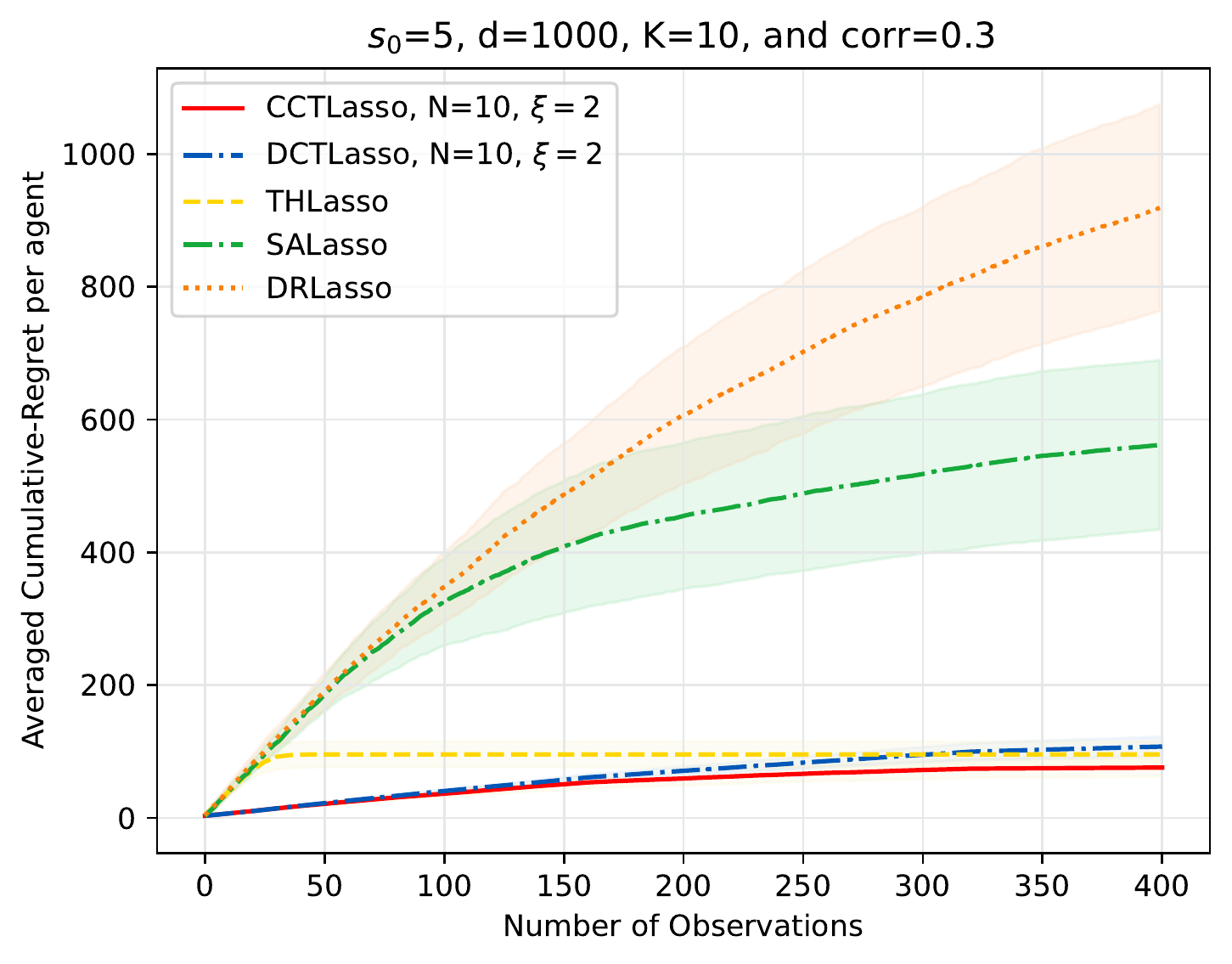}
    \includegraphics[width=0.4\linewidth]{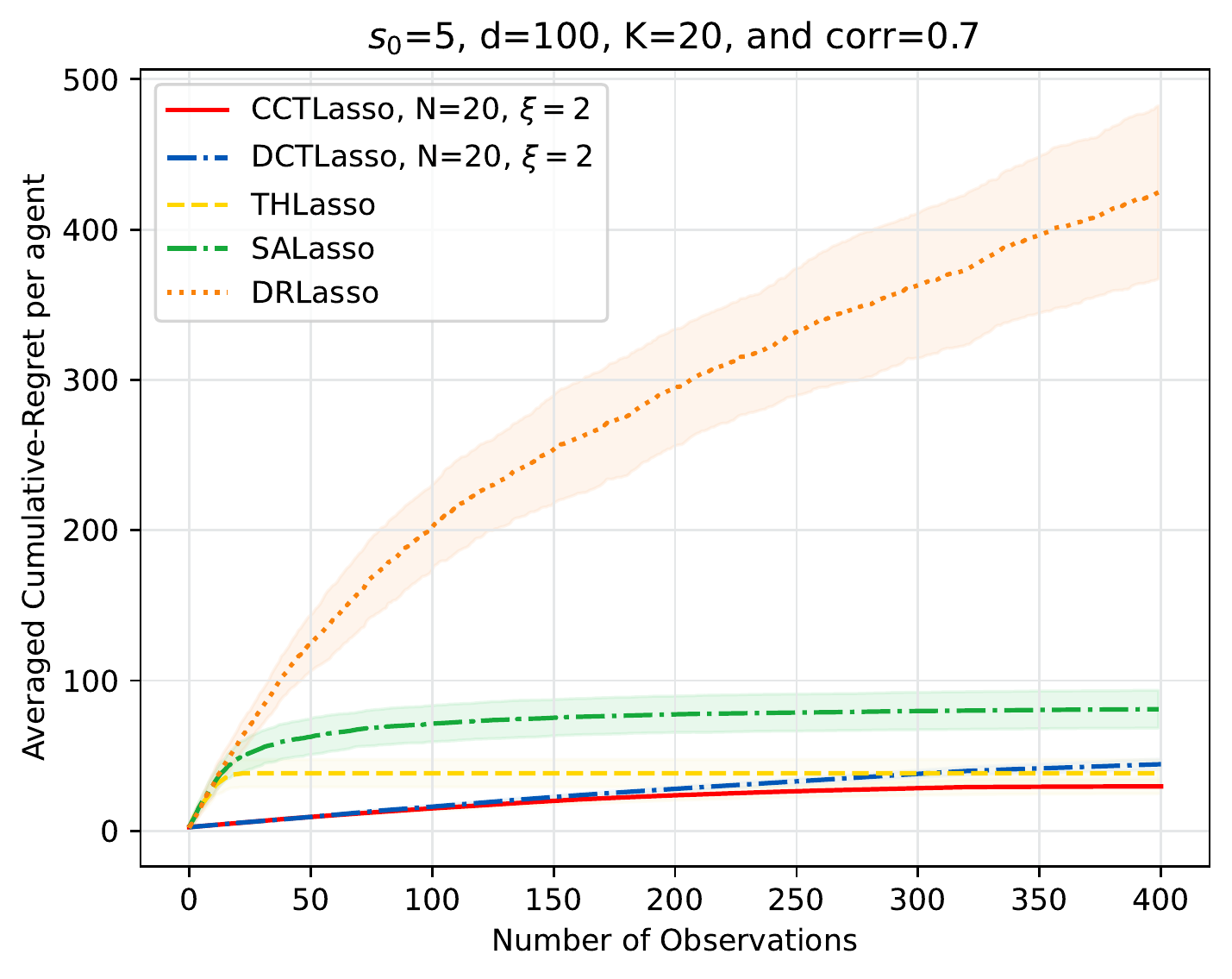}
    \includegraphics[width=0.4\linewidth]{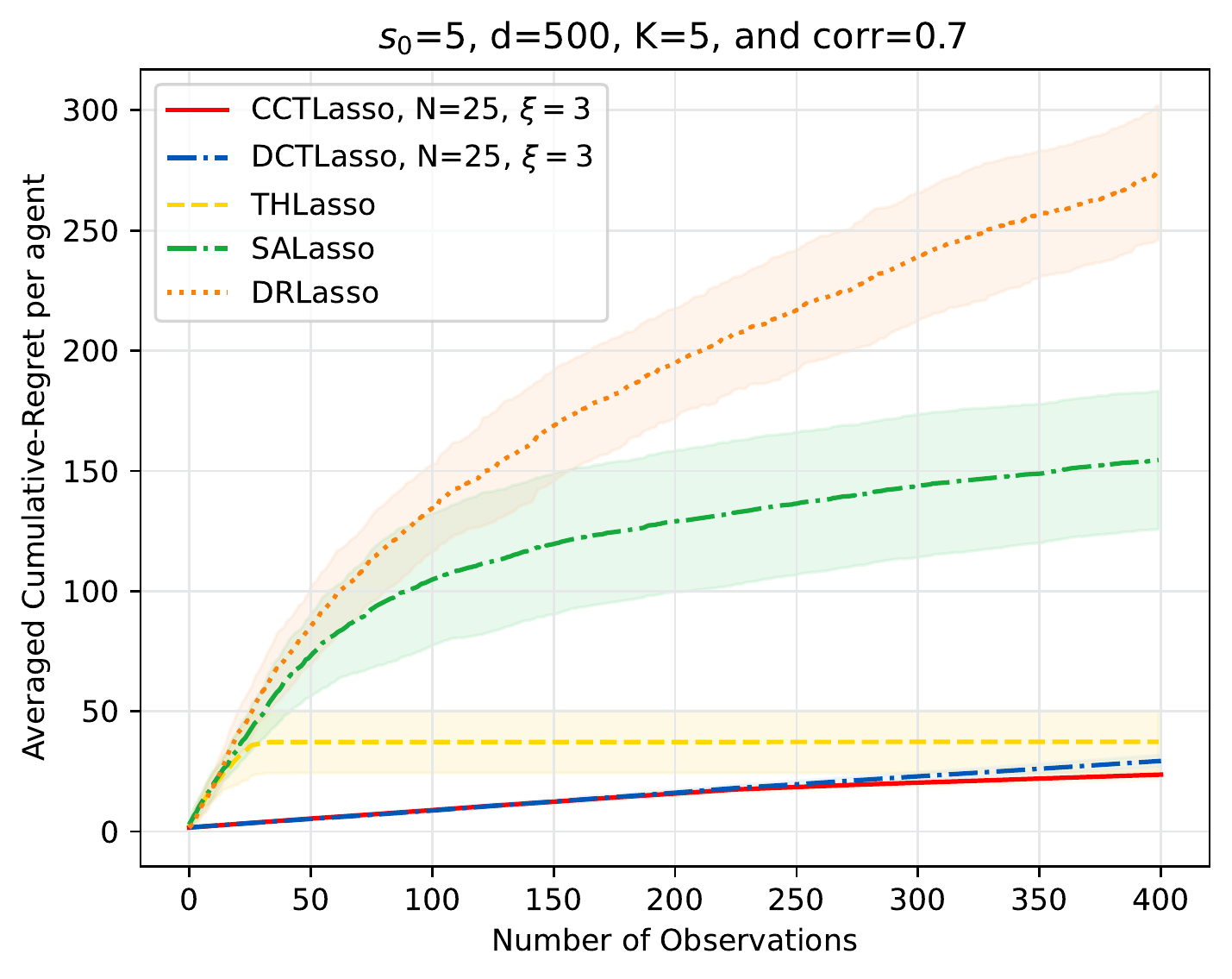}
    \includegraphics[width=0.4\linewidth]{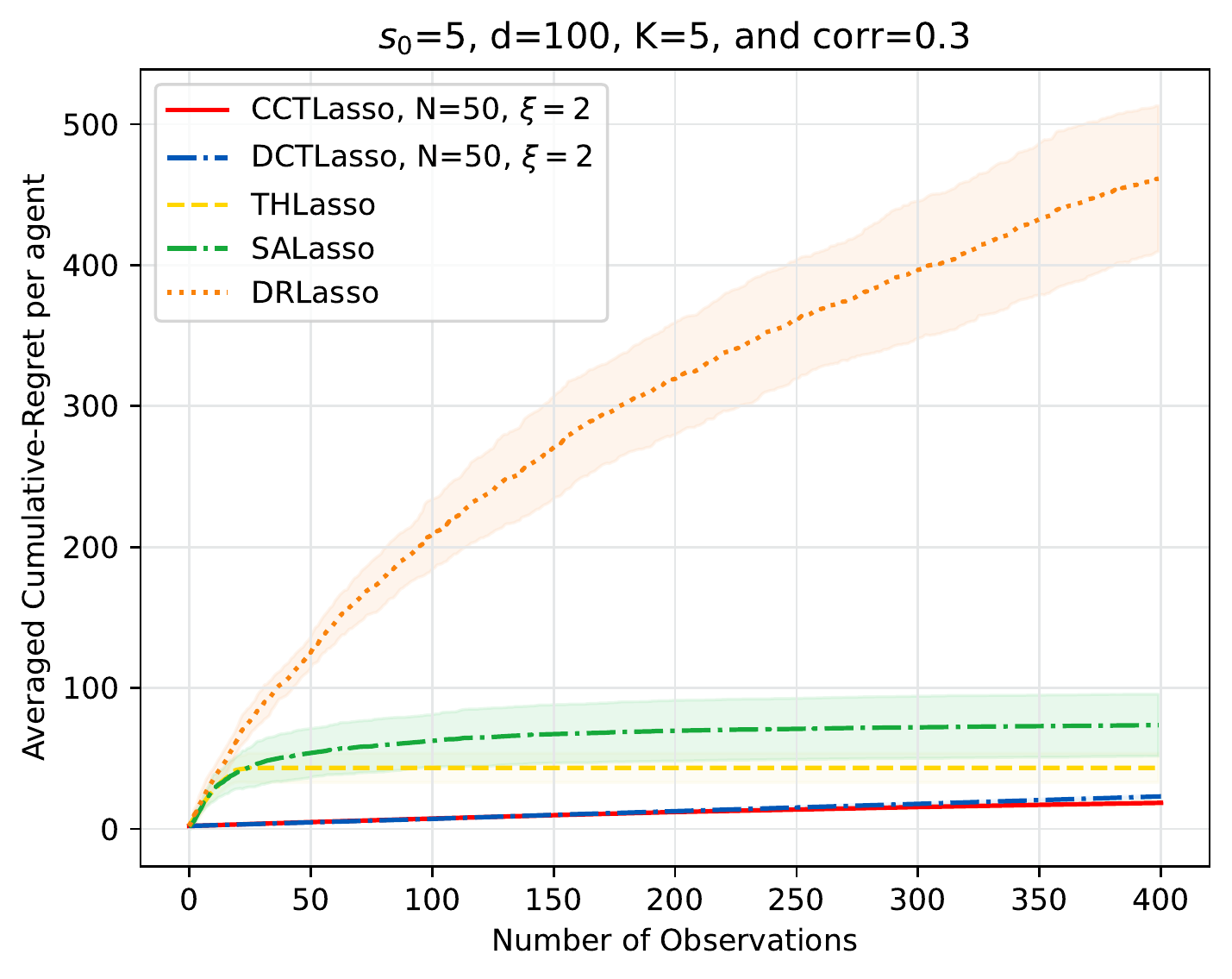}
    \caption{\textbf{Synthetic Data:} Comparison of CCTL and DCTL algorithms with state-of-the-art single-agent sparse linear bandit algorithms. The x-axis represents the number of observations per agent.}
    \label{fig:SvM}
\end{figure*}
\section{Experimental Results} 
\label{sec:expr}
In this section, we evaluate our methods described in Section \ref{sec:algorithm} in the context of solving a sparse linear bandit problem. Our theoretical analysis, as outlined in \ref{theo:CenRegret} and \ref{theo:deRegret}, demonstrates regret of order $\mathcal{O}(s_0 \log d + s_0 \sqrt{T})$ which is comparable to the state-of-the-art lasso-bandit algorithms. To evaluate our approach numerically, we conduct comparative experiments using both synthetic and real-world data. 

\subsection{Synthetic Data}
We focus on scenarios with $\theta^* \in \mathbb{R}^d$ is $s_0$-sparse. Specifically, we generate each non-zero element of $\theta^*$ in an i.i.d. fashion using a uniform distribution on the interval $[0.5, 2]$. Notably, parameter $\theta^*$ is the same for all agents. Given that, every component of the context distribution is endowed with a bounded density, Assumption \ref{asm:par_cons} holds. For each round $t$ and every agent $i$, we create $\mathcal{A}_t^i$ by sampling from a Gaussian distribution with mean zero and covariance matrix $V$. Here, for every $j$, $V_{j,j} = 1$ and for every $j \neq k$, $V_{j,k} = \rho^2$. We then normalize each $A_{t,k}^i$ such that its infinity-norm is at most $s_A = 5$ for all $k \in [K]$. Importantly, the feature vector components correlate over $[d]$ and $[K]$, and the Gram matrix's minimum eigenvalue is bounded below by a constant. Consequently, Assumptions \ref{asm:comp_cond} and \ref{asm:pos-def} hold. Additionally, the symmetry of the distribution confirms Assumption \ref{asm:relax_sym}. When the distribution is independent over arms, Proposition 1 in \cite{oh2021sparsity} confirms Assumption \ref{asm:balanced_cov}. It is worth noting that all agents share the $\mathcal{N}(0_K, V)$ distribution. Moreover, the additive noise is Gaussian, with i.i.d. realizations over rounds: $\omega_{t}^{i} \sim \mathcal{N}(0, 0.05)$. Furthermore, in the DCTL bandit algorithm, agents communicate through a network, which we model by a random connected graph $G = (N, E)$, by selecting the number of edges $|E|$ uniformly between $N-1$ and $2\times N$.




\subsubsection{Compare with Single-Agent Algorithms}
To evaluate the effectiveness of the CCTL and DCTL bandit algorithms, we firstly compare their performance against several single-agent algorithms, including the TH Lasso bandit \cite{ariu2022thresholded}, SA Lasso bandit \cite{oh2021sparsity}, and DR Lasso bandit \cite{kim2019doubly}. We fine-tune the hyper-parameter $\lambda_0$ in the range of [0.01, 0.5] for the CCTL bandit, DCTL bandit, SA Lasso bandit, and TH Lasso bandit algorithms to optimize their performance, while for DR Lasso bandit, we utilize the hyper-parameters provided in their respective code implementations. We conduct experiments by varying the values of $K$, $d$, $s_0$, and $\rho^2$, and report the results over $10$ instances for each experimental setting. The averaged cumulative regret per agent is presented in Figure \ref{fig:SvM}. Our results demonstrate that DCTL and CCTL bandit algorithms outperform the other algorithms in all scenarios, with the centralized approach performing slightly better. This finding aligns with the theoretical analysis that suggest the performance of the decentralized and centralized versions of cooperative thresholded Lasso are similar.
\begin{figure*}[!t]
    \centering
    \includegraphics[width=0.38\textwidth]{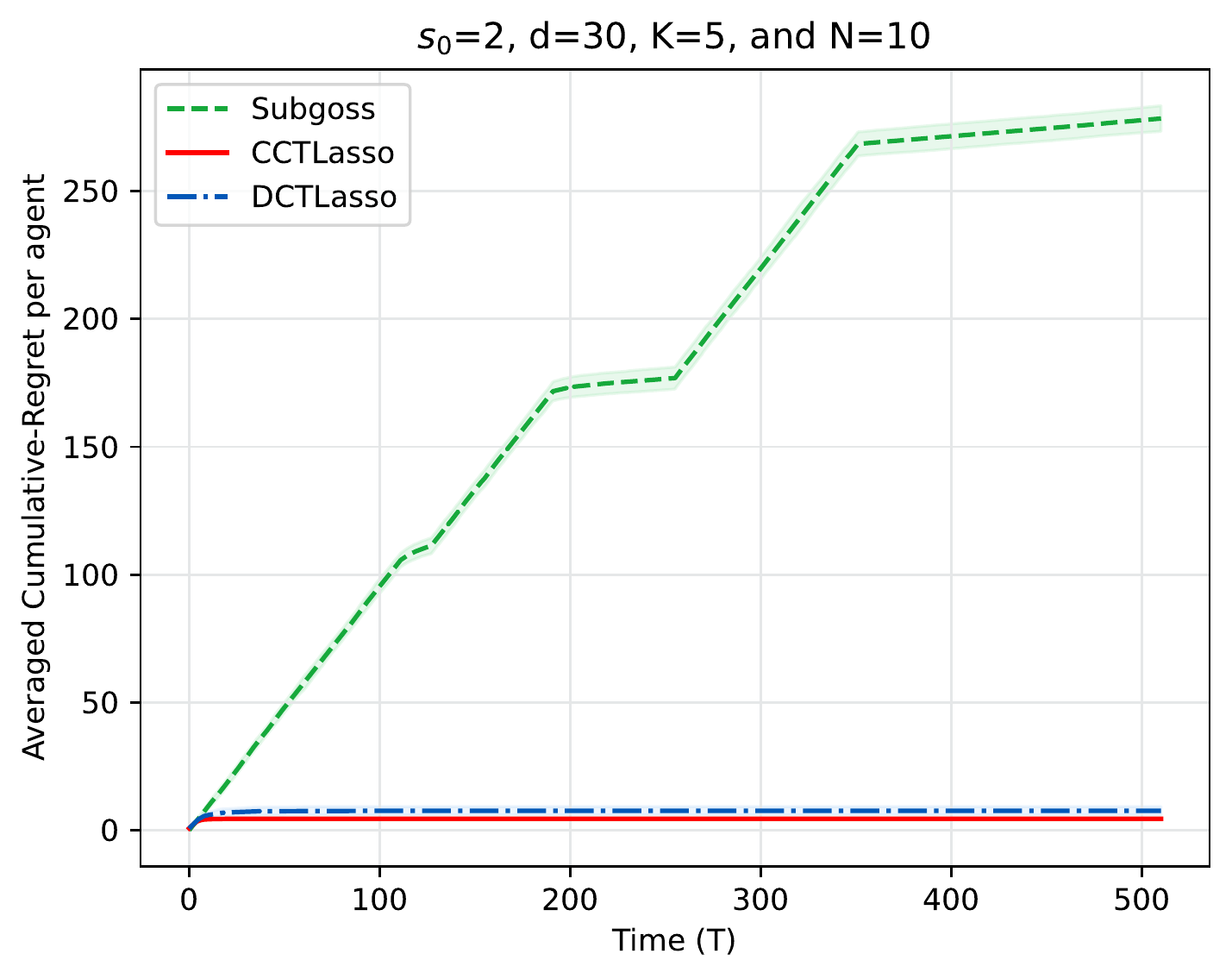}
    \includegraphics[width=0.38\textwidth]{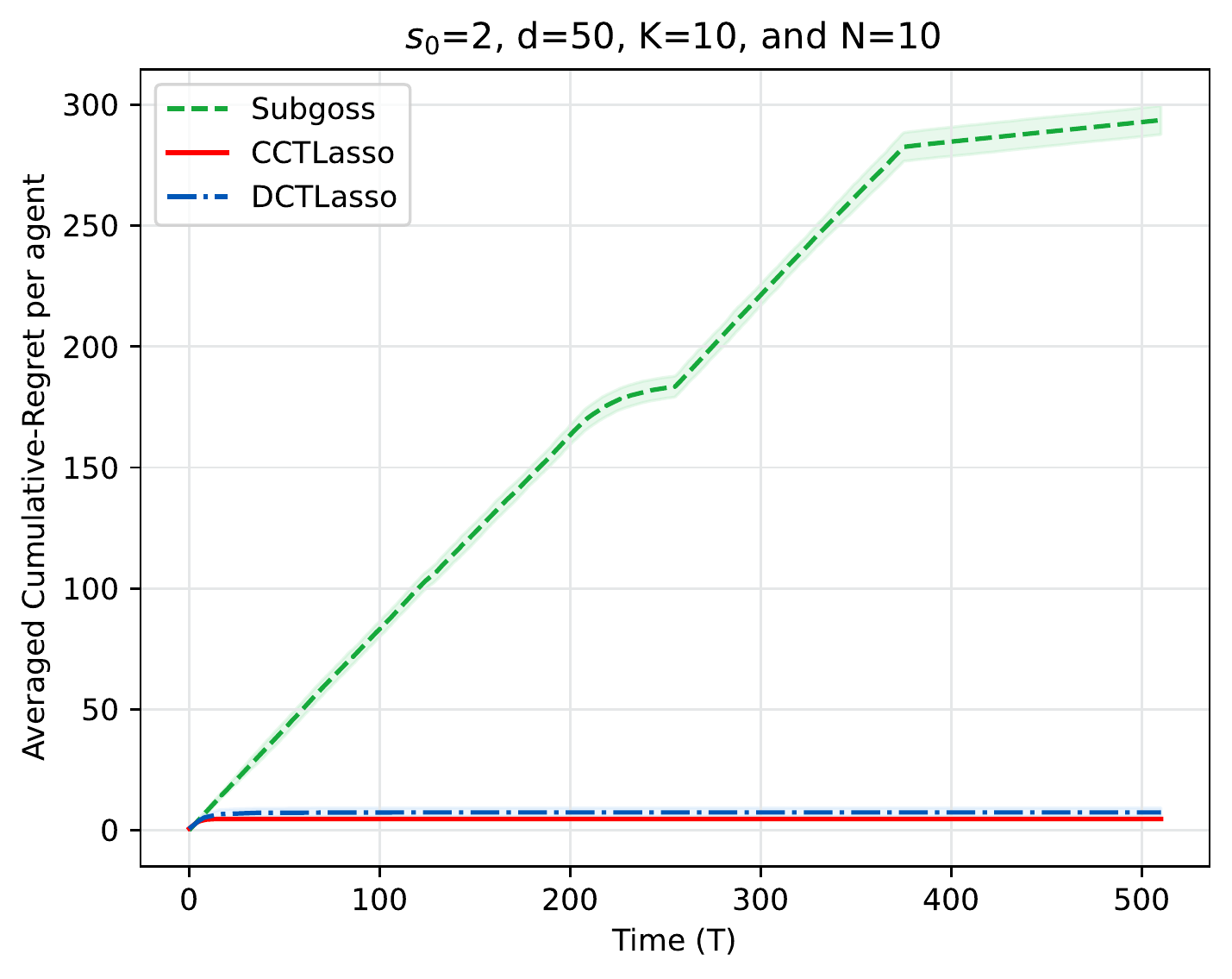}
    \caption{Comparison of CCTL and DCTL algorithms with Subgoss\cite{9786669}. The x-axis shows the number of observations per agent.}
    \label{fig:MvM}
\end{figure*}
\subsubsection{Compare with Multi-Agent Algorithm}
We compare our proposed method with the multi-agent low dimensional Linear Bandit method, namely, SubGoss \cite{9786669}. It assumes that an unknown parameter $\theta^*$ lies in one of many low-dimensional subspaces. Agents identify a small active set of subspaces and play actions only within this set, using pure exploration to identify the most likely subspace and then playing a projected version of the LinUCB algorithm to minimize regret within that subspace. The active set of subspaces is updated through collaboration and communication among agents. The algorithm has two phases in which the active subspaces remain fixed. In contrast to our problem setting, in this method, the agents have a collection of $K$ disjoint $m$-dimensional subspaces, and one contains the unknown parameter $\theta^*$. Despite the availability of such side information, our proposed method outperforms the SubGos algorithm, as demonstrated by the results of our experiments, presented in Figure \ref{fig:MvM}.

\subsection{Real-World Data}
\begin{figure}
    \centering
    \includegraphics[width=0.38\textwidth]{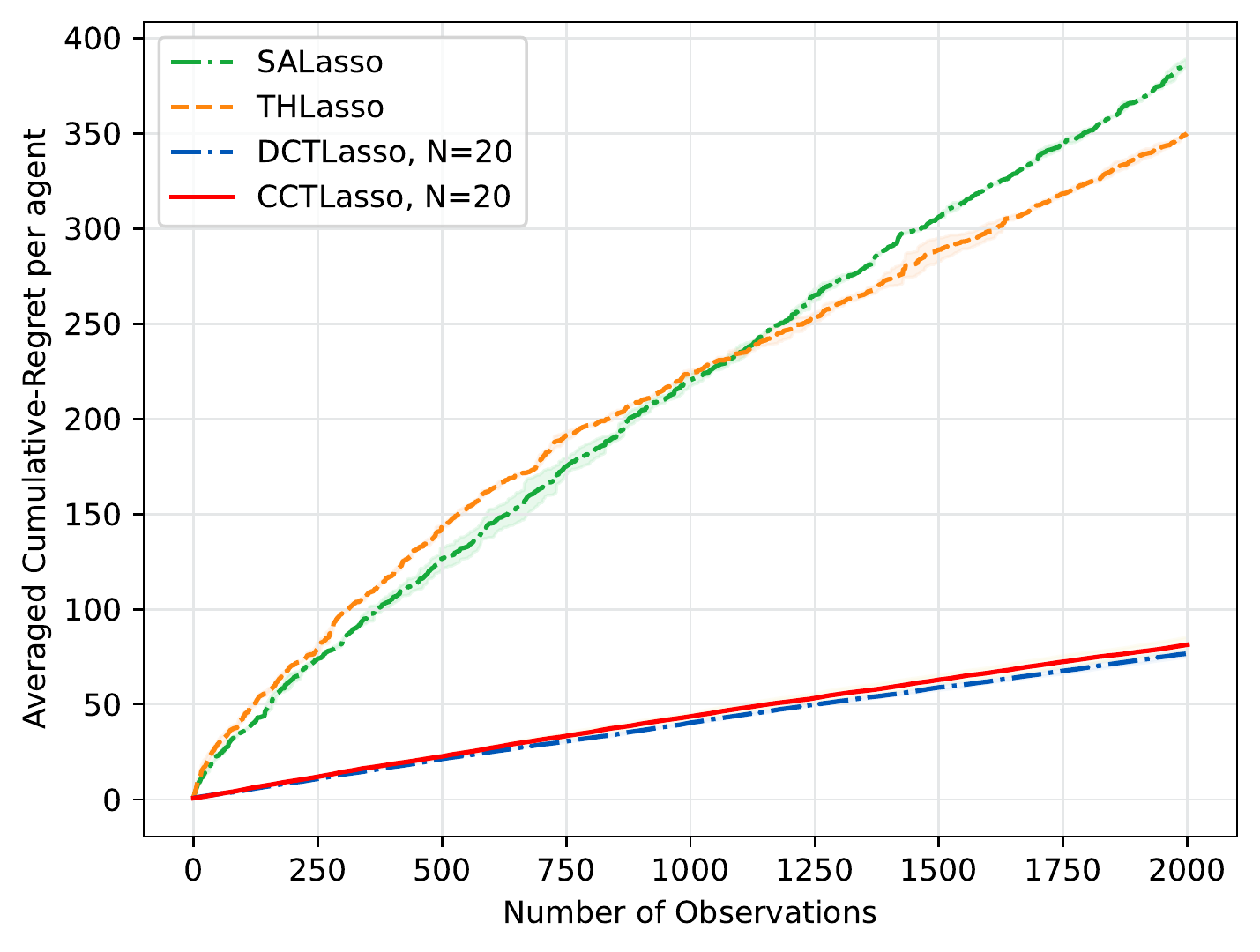}
    \caption{\textbf{Real Data:} Comparison of CCTL and DCTL algorithms with state-of-the-art single-agent sparse linear bandit algorithms. The x-axis represents the number of observations per agent.}
    \label{fig:real}
\end{figure}

In this section, we demonstrate the applicability of our method on real-world datasets. We utilize Movielens 1M dataset\footnote{Data is available at https://grouplens.org/datasets/movielens/1m/}, which contains approximately one million anonymous ratings from 6,000 users for 4,000 movies. We employ an SVD transformation with a dimensionality of $d = 70$. In each round, for each agent, we randomly suggest $K = 30$ movies. Agents use a bandit algorithm to select a movie (arm), aiming to choose the best one from $30$ choices that satisfy the general preferences of users. Figure \ref{fig:real} displays the results of the SA Lasso, TH Lasso, CCTL, and DCTL bandit algorithms. It is evident that the CCTL and DCTL bandit algorithms performed well in comparison to the other approaches. As discussed in the previous section, SubGoss has several limitations and its performance is not adequate. Therefore, we do not include it in the current comparison.

\section{Conclusion} \label{sec:conc}
In this paper, we introduce a method for solving the multi-agent sparse contextual linear bandit problem. Our approach leverages Lasso regression to reduce the problem's dimensions and utilizes the network structure to enable each agent to independently estimate the key dimensions and share this knowledge with others in only logarithmic time steps. Notably, our algorithm is the first to tackle row-wise distributed data in sparse linear bandits and delivers performance comparable to state-of-the-art single and multi-agent methods. This method has broad applicability for high-dimensional multi-agent problems, where efficient feature extraction is crucial for minimizing regret. Furthermore, we demonstrate that our proposed method achieves the same regret bound as \cite{ariu2022thresholded} approach while only performing dimension reduction in logarithmic time steps and a single thresholding stage, as opposed to the approach proposed in \cite{ariu2022thresholded} which performs dimension reduction in every time step. However, our theoretical analysis is limited by the way in which the threshold is defined. Our experimental results indicate that performance is not affected significantly by selecting all non-zero dimensions, whereas our theoretical approach requires a threshold to recover dimensions. Future research could explore ways to improve the theoretical framework to remove the dependency on the threshold value.

\bibliography{ecai}

\begin{thebibliography}{10}

\bibitem{abbasi2011improved}
Yasin Abbasi-Yadkori, D{\'a}vid P{\'a}l, and Csaba Szepesv{\'a}ri, `Improved
  algorithms for linear stochastic bandits', {\em Advances in neural
  information processing systems}, {\bf 24}, (2011).

\bibitem{pmlr-v22-abbasi-yadkori12}
Yasin Abbasi-Yadkori, David Pal, and Csaba Szepesvari,
  `Online-to-confidence-set conversions and application to sparse stochastic
  bandits', in {\em Proceedings of the Fifteenth International Conference on
  Artificial Intelligence and Statistics}, volume~22 of {\em Proceedings of
  Machine Learning Research}, pp. 1--9. PMLR, (21--23 Apr 2012).

\bibitem{Amani_Thrampoulidis_2021}
Sanae Amani and Christos Thrampoulidis, `Decentralized multi-agent linear
  bandits with safety constraints', {\em Proceedings of the AAAI Conference on
  Artificial Intelligence}, {\bf 35}(8),  6627--6635, (May 2021).

\bibitem{ariu2022thresholded}
Kaito Ariu, Kenshi Abe, and Alexandre Prouti{\`e}re, `Thresholded lasso
  bandit', in {\em International Conference on Machine Learning}, pp. 878--928.
  PMLR, (2022).

\bibitem{b975d286bb2d4217849b89298c4b6b0c}
Peter Auer, `Using confidence bounds for exploitation-exploration trade-offs',
  {\em Journal of machine learning research (JMLR)},  397--422, (2002).

\bibitem{buhlmann2011statistics}
Van De~Geer B{\"u}hlmann, {\em Statistics for high-dimensional data: methods,
  theory and applications}, Springer Science \& Business Media, 2011.

\bibitem{cella2021multi}
Leonardo Cella and Massimiliano Pontil, `Multi-task and meta-learning with
  sparse linear bandits', in {\em Uncertainty in Artificial Intelligence}, pp.
  1692--1702. PMLR, (2021).

\bibitem{ijcai2017p24}
Mithun Chakraborty, Kai Yee~Phoebe Chua, Sanmay Das, and Brendan Juba,
  `Coordinated versus decentralized exploration in multi-agent multi-armed
  bandits', in {\em Proceedings of the Twenty-Sixth International Joint
  Conference on Artificial Intelligence, {IJCAI-17}}, pp. 164--170, (2017).

\bibitem{pmlr-v108-chawla20a}
Ronshee Chawla, Abishek Sankararaman, Ayalvadi Ganesh, and Sanjay Shakkottai,
  `The gossiping insert-eliminate algorithm for multi-agent bandits', in {\em
  Proceedings of the Twenty Third International Conference on Artificial
  Intelligence and Statistics}, volume 108 of {\em Proceedings of Machine
  Learning Research}, pp. 3471--3481. PMLR, (26--28 Aug 2020).

\bibitem{9786669}
Ronshee Chawla, Abishek Sankararaman, and Sanjay Shakkottai, `Multi-agent
  low-dimensional linear bandits', {\em IEEE Transactions on Automatic
  Control},  1--1, (2022).

\bibitem{Degenne2020GamificationOP}
R{\'e}my Degenne, Pierre M'enard, Xuedong Shang, and Michal Valko,
  `Gamification of pure exploration for linear bandits', in {\em International
  Conference on Machine Learning}, (2020).

\bibitem{10.5555/3524938.3525195}
Abhimanyu Dubey and Alex Pentland, `Kernel methods for cooperative multi-agent
  contextual bandits', in {\em Proceedings of the 37th International Conference
  on Machine Learning}, ICML'20. JMLR.org, (2020).

\bibitem{pmlr-v119-dubey20a}
Abhimanyu Dubey and Alex~`Sandy' Pentland, `Cooperative multi-agent bandits
  with heavy tails', in {\em Proceedings of the 37th International Conference
  on Machine Learning}, volume 119 of {\em Proceedings of Machine Learning
  Research}, pp. 2730--2739. PMLR, (13--18 Jul 2020).

\bibitem{dubey2020differentially}
Abhimanyu Dubey and AlexSandy' Pentland, `Differentially-private federated
  linear bandits', {\em Advances in Neural Information Processing Systems},
  {\bf 33},  6003--6014, (2020).

\bibitem{210608902}
Avishek Ghosh, Abishek Sankararaman, and Kannan Ramchandran.
\newblock Adaptive clustering and personalization in multi-agent stochastic
  linear bandits, 2021.

\bibitem{gilton2017sparse}
Davis Gilton and Rebecca Willett, `Sparse linear contextual bandits via
  relevance vector machines', in {\em 2017 International Conference on Sampling
  Theory and Applications (SampTA)}, pp. 518--522. IEEE, (2017).

\bibitem{NEURIPS2020_7212a656}
Yassir Jedra and Alexandre Proutiere, `Optimal best-arm identification in
  linear bandits', in {\em Advances in Neural Information Processing Systems},
  volume~33, pp. 10007--10017. Curran Associates, Inc., (2020).

\bibitem{kim2019doubly}
Gi-Soo Kim and Myunghee~Cho Paik, `Doubly-robust lasso bandit', {\em Advances
  in Neural Information Processing Systems}, {\bf 32}, (2019).

\bibitem{1604.07706}
Nathan Korda, Bal\'{a}zs Sz\"{o}r\'{e}nyi, and Shuai Li, `Distributed
  clustering of linear bandits in peer to peer networks', in {\em Proceedings
  of the 33rd International Conference on International Conference on Machine
  Learning - Volume 48}, ICML'16, p. 1301–1309. JMLR.org, (2016).

\bibitem{lattimore2020bandit}
T.~Lattimore and C.~Szepesv{\'a}ri, {\em Bandit Algorithms}, Cambridge
  University Press, 2020.

\bibitem{10.5555/3305890.3305895}
Lihong Li, Yu~Lu, and Dengyong Zhou, `Provably optimal algorithms for
  generalized linear contextual bandits', in {\em Proceedings of the 34th
  International Conference on Machine Learning - Volume 70}, ICML'17, p.
  2071–2080. JMLR.org, (2017).

\bibitem{li2019improved}
Shuai Li, Wei Chen, Shuai Li, and Kwong-Sak Leung, `Improved algorithm on
  online clustering of bandits', in {\em Proceedings of the 28th International
  Joint Conference on Artificial Intelligence}, pp. 2923--2929, (2019).

\bibitem{7498076}
Setareh Maghsudi and Ekram Hossain, `Multi-armed bandits with application to 5g
  small cells', {\em IEEE Wireless Communications}, {\bf 23}(3),  64--73,
  (2016).

\bibitem{McMahan2016CommunicationEfficientLO}
H.~B. McMahan, Eider Moore, Daniel Ramage, Seth Hampson, and Blaise~Ag{\"u}era
  y~Arcas, `Communication-efficient learning of deep networks from
  decentralized data', in {\em International Conference on Artificial
  Intelligence and Statistics}, (2016).

\bibitem{oh2021sparsity}
Min-hwan Oh, Garud Iyengar, and Assaf Zeevi, `Sparsity-agnostic lasso bandit',
  in {\em International Conference on Machine Learning}, pp. 8271--8280. PMLR,
  (2021).

\bibitem{9079458}
Erick Schmidt, Nikolaos Gatsis, and David Akopian, `A gps spoofing detection
  and classification correlator-based technique using the lasso', {\em IEEE
  Transactions on Aerospace and Electronic Systems}, {\bf 56}(6),  4224--4237,
  (2020).

\bibitem{slivkins2011contextual}
Aleksandrs Slivkins, `Contextual bandits with similarity information', in {\em
  Proceedings of the 24th annual Conference On Learning Theory}, pp. 679--702.
  JMLR Workshop and Conference Proceedings, (2011).

\bibitem{JMLR:v19:17-025}
Sara van~de Geer, `On tight bounds for the lasso', {\em Journal of Machine
  Learning Research}, {\bf 19}(46),  1--48, (2018).

\bibitem{007.03812}
Daniel Vial, Sanjay Shakkottai, and R.~Srikant, `Robust multi-agent multi-armed
  bandits', in {\em Proceedings of the Twenty-Second International Symposium on
  Theory, Algorithmic Foundations, and Protocol Design for Mobile Networks and
  Mobile Computing}, MobiHoc '21, p. 161–170. Association for Computing
  Machinery, (2021).

\bibitem{1904.06309}
Yuanhao Wang, Jiachen Hu, Xiaoyu Chen, and Liwei Wang, `Distributed bandit
  learning: Near-optimal regret with efficient communication', {\em arXiv
  preprint arXiv:1904.06309}, (2019).

\bibitem{10.1145/3571306.3571432}
Jingren Wei and Shaileshh Bojja~Venkatakrishnan, `Decvi: Adaptive video
  conferencing on open peer-to-peer networks', in {\em Proceedings of the 24th
  International Conference on Distributed Computing and Networking}, ICDCN '23,
  p. 336–341. Association for Computing Machinery, (2023).

\bibitem{b2bcfa5d95c349b4b79272d770ab1a7e}
Marco Wiering, `Multi-agent reinforcement learning for traffic light control',
  in {\em ICML}, pp. 1151--1158, (2000).

\bibitem{Zhou2010ThresholdedLF}
Shuheng Zhou, `Thresholded lasso for high dimensional variable selection and
  statistical estimation', {\em arXiv: Statistics Theory}, (2010).

\end{thebibliography}

\onecolumn
\appendix

\section{Proof of Lemmas}
\subsection{Proof of Lemma \ref{lem:cen-supp}} 
\label{app:prf-supp}
Let $\hat{\theta}_t^i$ denote the Lasso estimator for agent $i$ during time step $t$, and define $v_t^i = \hat{\theta}_t^i - \theta^*$. We first state two additional Lemmas for a comprehensive assessment of the initial Lasso estimate's performance.
\begin{lemma}(Lemma F.1 in \cite{ariu2022thresholded}, for general choice of $\lambda_t$) 
\label{lem:eventG}
    Let us consider $\hat{\Sigma}_t^i := \frac{\sum_{s=1}^t A_s^i (A_s^i)^\top}{t}$, which represents the empirical covariance matrix derived from the context vectors that agent $i$ has selected until time step $t$. We further assume that $\hat{\Sigma}_t^i$ satisfies the compatibility condition \ref{asm:comp_cond} with the support $S(\theta^*)$ and the compatibility constant $\phi_{t,i}$. Given assumption \ref{asm:par_cons}, we have:
	\begin{align*}
		\forall i \in [N]: \mathbb{P}\left(\norm{v_t^i}_1 \leq \frac{4s_0\lambda_t}{\phi_{t,i}^2}\right) \geq 1- 2\exp(-\frac{t\lambda_t^2}{32\sigma^2s_A^2}+\log d).
	\end{align*}
\end{lemma}
\begin{lemma}(Lemma F.2 in \cite{ariu2022thresholded}) 
\label{lem:com-par}
	Let $C_0 := \min \{\frac{1}{2}, \frac{\phi_0^2}{512s_0s_A^2\nu C_b}\}$. For all $t \geq \frac{2 \log(2d^2)}{C_0^2}$ and for each agent $i$, we have
	\begin{align*}
		\mathbb{P}\left(\phi^2\big(\hat{\Sigma}_t^i, S(\theta^*)\big) \geq \frac{\phi_0^2}{4\nu C_b}\right) \geq 1-\exp(-\frac{tC_0^2}{2}).
	\end{align*}
\end{lemma}
This lemma establishes that the discrepancy between the compatibility constant of $\hat{\Sigma}_t^i$ and that of $\Sigma$ is relatively small. We shall now proceed with the proof, following the method described in \cite{Zhou2010ThresholdedLF}. To begin with, we define the event $\mathcal{G}_t^i$ for each agent $i \in [N]$ as follows:
\begin{align*}
	\mathcal{G}_t^i := \left\{ \norm{v_t^i}_1 \leq \frac{4s_0\lambda_t}{\phi_{t,i}^2} \right\}.
\end{align*}
Subsequently, we assume that $\mathcal{G}_t^i$ holds. On this premise, we can base our arguments on the following:
\begin{align*}
	\norm{v_t^i}_1 \geq \norm{v_{t,S(\theta^*)^c}^i}_1 &= \sum_{j \in S(\theta^*)^c}|(\hat{\theta}_t^i)_j| \notag \\
	& \geq \sum_{j \in S(\theta^*)^c \cap \hat{S}_t^i} |(\hat{\theta}_t^i)_j| \notag \\
	& = \sum_{j \in \hat{S}_t^i \setminus S(\theta^*)} |(\hat{\theta}_t^i)_j| \notag \\
	& \geq |\hat{S}_t^i \setminus S(\theta^*)| \times \mathcal{T}_t,
\end{align*}
where $S(\theta^*)^c := [d] \setminus S(\theta^*)$ and $\mathcal{T}_t$ is a monotonically decreasing function of $t$ that reflects the threshold for each dimension reduction step. Then for each agent $i \in [N]$ and $t\in [T]$ the following holds
\begin{align}
\label{ag:sz:bnd}
	|\hat{S}_t^i \setminus S(\theta^*)| \leq \frac{\norm{v_t^i}_1}{\mathcal{T}_t} \leq \frac{4 s_0}{\phi_{t,i}^2 N}.
\end{align}
Besides, $\forall j \in S(\theta^*)$,
\begin{align*}
	|(\hat{\theta}_t^i)_j| &\geq \theta^*_{\min} - \norm{v_{t, S(\theta^*)}^i}_{\infty} \notag \\
	&\geq \theta^*_{\min} - \norm{v_{t, S(\theta^*)}^i}_1 \notag \\
	&\geq \theta^*_{\min} - \frac{4s_0\lambda_t}{\phi_{t,i}^2}.
\end{align*}

Therefore, when $t$ is large enough so that $\theta^*_{\min} - \frac{4s_0\lambda_t}{\phi_{t,i}^2} \ge \mathcal{T}_t$, we have $S(\theta^*) \subset \hat{S}_t^i$. Putting this result and \eqref{ag:sz:bnd} together using Lemma \ref{lem:eventG}, if $\theta^*_{\min} \ge \frac{4s_0\lambda_t}{\phi_{t,i}^2}+\mathcal{T}_t$ we can conclude the following probability for $\hat{S}_t=\underset{i=1}{\overset N{\bigcup}} \hat{S}_t^i$:
\begin{align*}
	\mathbb{P}\left(|\hat{S}_t \setminus S(\theta^*)| \le \frac{4 s_0}{\phi_{t,i}^2} \text{ and } S(\theta^*) \subset \hat{S}_t\right) \ge 1 -  \left(2\exp(-\frac{t\lambda_t^2}{32\sigma^2s_A^2}+\log d)\right)^{N}.
\end{align*}
Finally, according to Lemma \ref{lem:com-par}, for each agent $i$ we substitute $\phi_{t,i}^2$ by  $\phi_0^2 / (4 \nu C_b)$. So, if $\theta^*_{\min} \ge \frac{16 s_0 \nu C_b \lambda_t}{\phi_0^2} + \mathcal{T}_t$ we get
\begin{align*}
	\mathbb{P}\left(|\hat{S}_t \setminus S(\theta^*)| \le \frac{16 s_0 \nu C_b}{\phi_0^2} \text{ and } S(\theta^*) \subset \hat{S}_t\right) \ge 1 -  \left(2\exp(-\frac{t\lambda_t^2}{32\sigma^2s_A^2}+\log d)\right)^N - \exp(-\frac{NtC_0^2}{2}).
\end{align*}
As we reduce the dimension only in $t' = \xi^{\lfloor \log_\xi t \rfloor}$, for all $t \in [T]$ we have
\begin{align*}
	\mathbb{P}\left(|\hat{S}_t \setminus S(\theta^*)| \le \frac{16 s_0 \nu C_b}{\phi_0^2} \text{ and } S(\theta^*) \subset \hat{S}_t\right) \ge 1 - \left(2\exp(-\frac{t'\lambda_{t'}^2}{32\sigma^2s_A^2}+\log d)\right)^N - \exp(-\frac{Nt'C_0^2}{2}).
\end{align*}
\qed
\subsection{Proof of Lemma \ref{lem:inst-rgt}} 
\label{app:prf-inst-rgt}
The instantaneous expected regret of agent $i$ at round $t$ is defined as
\begin{align*}
	r_t^i = \mathbb{E}\big[\max_{A \in \mathcal{A}_t^i}\langle A - A_t^i, \theta^* \rangle\big].
\end{align*}
Define event $\mathcal{R}_k^i := \left\{\mathcal{A}_t^i \in \mathbb{R}^{K \times d}: k \in \arg \max_{k'}\langle A_{t,k'}^i, \theta^* \rangle \right\}$ and $\mathcal{G}_{t,i}^{\lambda} := \left\{\lambda_{\min}(\hat{\Sigma}_{\hat{S}_t}^i) \geq \lambda \right\}$. Then, we can bound $r_t^i$ as follows
\begin{align}\nonumber
	r_t^i & \stackrel{(a)}{\le} \sum_{k=1}^K\mathbb{E}\left[r_t^i|\mathcal{A}_t^i \in \mathcal{R}_k^i\right] \times \mathbb{P}\big(\mathcal{A}_t^i \in \mathcal{R}_k^i\big) \\ \nonumber
	&= \sum_{k=1}^K\mathbb{E}\left[\langle A_{t,k}^i - A_t^i, \theta^* \rangle |\mathcal{A}_t^i \in \mathcal{R}_k^i\right] \times \mathbb{P} \big(\mathcal{A}_t^i \in \mathcal{R}_k^i\big) \\ \nonumber
	&\leq \sum_{k=1}^K\mathbb{E}\left[\langle A_{t,k}^i - A_t^i, \theta^* \rangle |\mathcal{A}_t^i \in \mathcal{R}_k^i \cap \mathcal{E}_t \cap \mathcal{G}_{t,i}^{\frac{\alpha}{4 \nu C_b}}\right] \times 1 + 2 K s_A s_1 \mathbb{P}\big(\mathcal{A}_t^i \in \mathcal{R}_k^i \cap ((\mathcal{E}_t)^c \cup (\mathcal{G}_{t,i}^{\frac{\alpha}{4 \nu C_b}})^c)\big) \\
	&\stackrel{(b)}{\le} \sum_{k=1}^K\mathbb{E}\left[\langle A_{t,k}^i - A_t^i, \theta^* \rangle |\mathcal{A}_t^i \in \mathcal{R}_k^i \cap \mathcal{E}_t \cap \mathcal{G}_{t,i}^{\frac{\alpha}{4 \nu C_b}}\right] + 2 K s_A s_1 \big(\mathbb{P}((\mathcal{E}_t)^c) + \mathbb{P}((\mathcal{G}_{t,i}^{\frac{\alpha}{4 \nu C_b}})^c | \mathcal{E}_t)\big), \label{regret_ineq}
\end{align}
where inequality $(a)$ holds, because the events $\mathcal{R}_k^i$s may intersect. According to $\mathbb{P}(A \bigcup B) \le \mathbb{P}(A) + \mathbb{P}(B|A^c)$, we have inequality $(b)$.
	
Now, let's bound $\langle A_{t,k}^i - A_t^i, \theta^* \rangle$. For simplicity, we put $A_i^*$ instead of $A_{t,k}^i$.
\begin{align}
	\langle A^*_i - A_t^i, \theta^* \rangle &= \langle A^*_i, \theta^* \rangle - \langle A_t^i, \theta^* \rangle \notag \\
	&= \langle A^*_i, \theta^* \rangle - \langle A^*_i, \hat{\theta}_t^i \rangle + \langle A^*_i, \hat{\theta}_t^i\rangle  - \langle A_t^i, \theta^* \rangle  \notag \\
	&\leq \langle A^*_i, \theta^* \rangle - \langle A^*_i, \hat{\theta}_t^i \rangle + \langle A_t^i, \hat{\theta}_t^i\rangle  - \langle A_t^i, \theta^* \rangle  &&\text{since $A_t^i$ is the $\arg\max$} \notag \\
	&= \langle A^*_i, \theta^* - \hat{\theta}_t^i  \rangle +  \langle A_t^i, \hat{\theta}_t^i - \theta^* \rangle. \label{eq:first-part}
\end{align}
Let $\omega_i = (\omega_1^i,\omega_2^i, \ldots, \omega_t^i)^T \in \mathbb{R}^{t \times 1}$ and $A_i = (A_1^i, \ldots, A_t^i) \in \mathbb{R}^{t\times d}$. As a ridge estimator, we know that:
\begin{align*}
	\hat{\theta}_t^i &= (A_i^T A_i + I)^{-1} A_i^T (A_i\theta^* + \omega_i) \notag \\
	&= (A_i^T A_i + I)^{-1} A_i^T\omega_i + (A_i^TA_i + I)^{-1}(A_i^T A_i + I)\theta^* - (A_i^T A_i + I)^{-1}\theta^* \notag \\
	&= (A_i^T A_i + I)^{-1} A_i^T\omega_i + \theta^* - (A_i^T A_i + I)^{-1}\theta^*. \\
\end{align*}
Then, we can obtain $\hat{\theta}_t^i - \theta^* = (A_i^T A_i + I)^{-1} A_i^T\omega_i - (A_i^T A_i + I)^{-1}\theta^*$. For any arbitrary $B \in \mathbb{R}^d$, we have:
\begin{align*}
	\langle B, \hat{\theta}^i_t - \theta^* \rangle &= B^T (A_i^T A_i + I)^{-1} A_i^T\omega_i - B^T (A_i^T A_i + I)^{-1}\theta^* \\ 
	&= B^T(M_t^i)^{-1} A_i^T\omega_i - B^T(M_t^i)^{-1}\theta^* \notag \\
	&= \langle B, A_i^T\omega_i \rangle_{(M_t^i)^{-1}} - \langle B, \theta^* \rangle_{(M_t^i)^{-1}},
\end{align*}
where $M_t^i = A_i^T A_i + I$ and it is  positive definite. Using Cauthy-Schwarz inequality, we get \cite{abbasi2011improved}
\begin{align} \label{p1}
	|\langle B, \hat{\theta}^i_t - \theta^* \rangle| \leq \norm{B}_{(M_t^i)^{-1}}(\norm{A_i^T\omega_i}_{(M_t^i)^{-1}} + \norm{\theta^*}_2). 
\end{align}
For any $\delta > 0$, with probability at least $1-\delta$ we have \cite{abbasi2011improved}
\begin{align}
\label{p2}
	\forall t \geq 0, \forall i\in [N] \geq 0: \quad \norm{A_i^T\omega_i}_{(M_t^i)^{-1}} \leq \sigma \sqrt{2\log(\frac{\det(M_t^i)^{1/2}}{\delta})}.
\end{align}
Combining \eqref{p1} and \eqref{p2}, for all $t \ge 0$ and for each agent $i$ we obtain
\begin{gather*}
	|\langle B, \hat{\theta}^i_t - \theta^* \rangle| \leq \norm{B}_{(M_t^i)^{-1}} (\sigma \sqrt{2\log(\frac{\det(M_t^i)^{1/2}}{\delta})} + \norm{\theta^*}_2).
\end{gather*}
Using inequality \eqref{eq:first-part}, if put $B = A_i^* - A_t^i$, we have:
\begin{align}
\label{inprbnd}
	\langle A_i^* - A_t^i, \theta^* \rangle \leq (\norm{A_i^*}_{(M_t^i)^{-1}}+\norm{A_t^i}_{(M_t^i)^{-1}})(\sigma \sqrt{2\log(\frac{\det(M_t^i)^{1/2}}{\delta})} + \norm{\theta^*}_2).
\end{align}
Subsequently, by utilizing inequality \eqref{regret_ineq} and bounding the first term using \eqref{inprbnd}, we derive an upper bound for the instantaneous regret as
\begin{gather*}
			r_t^i \le \sum_{k=1}^K \mathbb{E} \left[(\norm{A_{t,k}^i}_{(M_t^i)^{-1}}+\norm{A_t^i}_{(M_t^i)^{-1}})(\sigma \sqrt{2 \log(\frac{\det(M_t^i)^{1/2}}{\delta})} + \norm{\theta}^*_2) |\mathcal{A}_t^i \in \mathcal{R}_k^i, \mathcal{E}_t, \mathcal{G}_{t,i}^{\frac{\alpha}{4 \nu C_b}}\right] \\
            + 2 K s_A s_1 \big(\mathbb{P}((\mathcal{E}_t)^c) + \mathbb{P}((\mathcal{G}_{t,i}^{\frac{\alpha}{4 \nu C_b}})^c | \mathcal{E}_t)\big).
		\end{gather*}
\qed
%
\section{Proof of Theorem \ref{theo:CenRegret}} 
\label{app:prv-CenRegret}
To derive the upper bound for the expected cumulative regret of an arbitrary agent $i$ until the time horizon $T$, we must sum up the instantaneous regret from $t=1$ up to $t=T$
\begin{align}
	R_i(T) &= \sum_{t=1}^T \mathbb{E}[\max_{A\in \mathcal{A}_t^i}\langle A - A_t^i, \theta^* \rangle] \notag \\ \nonumber
	&\leq 2s_As_1\tau + \sum_{t=\tau+1}^T r_t^i \\ \nonumber
	&\stackrel{(a)}{\le} 2s_As_1\tau  + \sum_{t=\tau+1}^T \sum_{k=1}^K\mathbb{E} \left[(\norm{A_{t,k}^i}_{(M_t^i)^{-1}}+\norm{A_t^i}_{(M_t^i)^{-1}})(\sigma \sqrt{2\log(\frac{\det(M_t^i)^{1/2}}{\delta})} + \norm{\theta^*}_2) |\mathcal{A}_t^i \in \mathcal{R}_k^i, \mathcal{E}_t, \mathcal{G}_{t,i}^{\frac{\alpha}{4 \nu C_b}}\right] \\ 
	&\quad + 2 K s_A s_1 (\mathbb{P}((\mathcal{E}_t)^c) + \mathbb{P}((\mathcal{G}_{t,i}^{\frac{\alpha}{4 \nu C_b}})^c | \mathcal{E}_t)), \label{first}
\end{align}
where inequality $(a)$ holds according to Lemma \ref{lem:inst-rgt}. We define $l_t^i := (\norm{A^*}_{(M_t^i)^{-1}}+\norm{A_t^i}_{(M_t^i)^{-1}})(\sigma \sqrt{2\log(\frac{\det(M_t^i)^{1/2}}{\delta})}+ \norm{\theta^*}_2)$, as presented in \cite{abbasi2011improved}. Subsequently, we assume that the dimension between the time steps $t=1$ and $t=\Upsilon$ remains fixed at $d_{\Upsilon}$. Therefore, we obtain
\begin{align*}
	L_\Upsilon := \sum_{t=1}^{\Upsilon} l_t^i &\leq \sqrt{\Upsilon\sum_{t=1}^{\Upsilon}(l_t^i)^2} \notag \\
	&\stackrel{(a)}{\le} \sqrt{\Upsilon\sum_{t=1}^{\Upsilon} [2\norm{A^*}_{(M_t^i)^{-1}}^2+2\norm{A_t^i}_{(M_t^i)^{-1}}^2] (4\sigma^2\log(\frac{\det(M_{\Upsilon}^i)^{1/2}}{\delta})+2\norm{\theta^*}_2^2)} \notag \\
	&\stackrel{(b)}{\le} 2\sqrt{\Upsilon} \sqrt{(C_a d_\Upsilon \log \Upsilon) (2\sigma^2(\log(\det(M_{\Upsilon}^i)^{1/2}) - \log \delta)+\norm{\theta^*}_2^2)} \\
	&= 2\sqrt{\Upsilon} \sqrt{(C_a d_\Upsilon \log \Upsilon) (2\sigma^2(\frac{\log(\det(M_{\Upsilon}^i))}{2} - \log \delta)+\norm{\theta^*}_2^2)} \\
	&\stackrel{(c)}{\le} 2\sqrt{\Upsilon} \sqrt{(C_a d_\Upsilon \log \Upsilon) (2\sigma^2(\frac{C_a d_\Upsilon \log \Upsilon + \log \det(M_1^i)}{2} - \log \delta)+\norm{\theta^*}_2^2)} \\
	&= 2\sqrt{\Upsilon} \sqrt{\sigma^2 C_a^2 d_\Upsilon^2 \log^2 \Upsilon + (\norm{\theta^*}_2^2 - 2 \sigma^2 \log \delta)(C_a d_\Upsilon \log \Upsilon)},
\end{align*}
where inequality $(a)$ holds according to $(a + b)^2 \le 2 a^2 + 2 b^2$, proof of Lemma 13 in \cite{b975d286bb2d4217849b89298c4b6b0c} justifies inequality $(b)$, where $C_a$ is a positive constant. Moreover, inequality $(c)$ is a result of Lemma 12 in \cite{abbasi2011improved}. From Lemma \ref{lem:cen-supp}, we know that $d_\Upsilon \leq 
 s_0 + \frac{16 s_0 \nu C_b}{\phi_0}$, then we obtain
\begin{align}
\label{fixdupbnd}
    L_\Upsilon \le 8\sqrt{\Upsilon} \sqrt{\sigma^2 C_a^2 (s_0 + \frac{16 s_0 \nu C_b}{\phi_0})^2 \log^2 \Upsilon+ (\norm{\theta^*}_2^2 - 2 \sigma^2 \log \delta) (s_0 + \frac{16 s_0 \nu C_b}{\phi_0}) C_a \log \Upsilon}.
\end{align}
Now, in our setting, we have
\begin{align*}
    \sum_{t=1}^T l_t^i &\leq \sum_{f=1}^{\lceil \log_\xi T \rceil} L_{\xi^f} \\
    &\stackrel{(a)}{\le} 8  \sqrt{\sigma^2 C_a^2 (s_0 + \frac{16 s_0 \nu C_b}{\phi_0})^2 \log^2 T+ (\norm{\theta^*}_2^2 - 2 \sigma^2 \log \delta) (s_0 + \frac{16 s_0 \nu C_b}{\phi_0}) C_a \log T} \sum_{f=1}^{\lceil \log_\xi T \rceil} \xi^{f/2} \\
    &\le \frac{8\sqrt{\xi}}{\sqrt{\xi} - 1} \sqrt{\sigma^2 C_a^2 (s_0 + \frac{16 s_0 \nu C_b}{\phi_0})^2 \log^2 T + (\norm{\theta^*}_2^2 - 2 \sigma^2 \log \delta) C_a (s_0 + \frac{16 s_0 \nu C_b}{\phi_0})\log T} (\sqrt{\xi T} - 1),
\end{align*}
where inequality $(a)$ holds according to \eqref{fixdupbnd}. Combining this result with inequality \eqref{first} we obtain 
\begin{align}
	R_i(T) 
	   &\le 2s_As_1\tau \notag \\ 
	   &\quad + \frac{8\sqrt{\xi}}{\sqrt{\xi} - 1} K \sqrt{\sigma^2 C_a^2 (s_0 + \frac{16 s_0 \nu C_b}{\phi_0})^2 \log^2 T + (\norm{\theta^*}_2^2 - 2 \sigma^2 \log \delta) C_a (s_0 + \frac{16 s_0 \nu C_b}{\phi_0})\log T} (\sqrt{\xi T} - 1) \notag \\ 
	   &\quad + 2 K s_A s_1 \sum_{t=\tau + 1}^T (\mathbb{P}((\mathcal{E}_t)^c) + \mathbb{P}((\mathcal{G}_{t,i}^{\frac{\alpha}{4 \nu C_b}})^c | \mathcal{E}_t)) \notag \\
	   &\stackrel{(a)}{\le} 2s_As_1\tau \notag \\ 
	   &\quad + \frac{8\sqrt{\xi}}{\sqrt{\xi} - 1} K \sqrt{\sigma^2 C_a^2 (s_0 + \frac{16 s_0 \nu C_b}{\phi_0})^2 \log^2 T + (\norm{\theta^*}_2^2 - 2 \sigma^2 \log \delta) C_a (s_0 + \frac{16 s_0 \nu C_b}{\phi_0})\log T} (\sqrt{\xi T} - 1) \notag \\
       &\quad + 2 K s_A s_1 \sum_{t=\tau + 1}^T \left( (2\exp(-\frac{t'\lambda_{t'}^2}{32\sigma^2s_A^2}+\log d)\right)^N + \exp(-\frac{Nt'C_0^2}{2}) + \mathbb{P}\big((\mathcal{G}_{t,i}^{\frac{\alpha}{4 \nu C_b}})^c | \mathcal{E}_t)\big), \label{second}
\end{align}
where Lemma \ref{lem:cen-supp} justifies inequality $(a)$. For bounding the first term $\sum_{t=\tau + 1}^T \left(2\exp(-\frac{t'\lambda_{t'}^2}{32\sigma^2s_A^2}+\log d)\right)^N$, we have
\begin{align*}
    \sum_{t=\tau+1}^T [2\exp(-\frac{t' \lambda_{t'}^2}{32 \sigma^2s_A^2}+ \log d)]^N 
    &= \sum_{t=\tau+1}^T 2^N \exp(N(-\frac{ \lambda_0^2 \log t' \log d}{16\sigma^2s_A^2}+ \log d)) \\
    &\stackrel{(a)}{=} \sum_{t=\tau+1}^T 2^N \exp(N(-c\log(\xi^{\lfloor \log_\xi t \rfloor})\log d+ \log d)) \\
    &\leq \sum_{t=\tau+1}^T 2^N \exp(N(c \log \xi \log d - c\log t \log d+ \log d)) \\
    &\stackrel{(b)}{\le} \sum_{t=\tau+1}^T 2^N \exp(N(-\frac{1}{2}c\log t \log d)) \\
    &\stackrel{(c)}{\le} \sum_{t=\tau+1}^T 2^N \exp(- 2N\log t) \\
    &= \sum_{t=\tau+1}^T \frac{2^N}{t^{2N}} \\
    &\leq \int_{t=\tau+1}^T \frac{2^N}{t^{2N}} dt \\
    &\stackrel{(d)}{\le} \int_{t = (\tau+1)/\sqrt{2}}^T \frac{1}{t^{2N}} dt \\
    &\leq \int_{t = 1}^T \frac{1}{t^{2N}} dt \leq \frac{1-T^{1-2N}}{2N-1},
\end{align*}
where for $(a)$, we perform a substitution by letting $\lambda_0 = 4\sqrt{c} \sigma s_A$. The assumption $\tau \ge \exp(2\log \xi + \frac{2}{c})$ justifies inequality $(b)$. Furthermore, we consider $(c)$ by assuming that $d \ge \exp(\frac{4}{c})$. Finally, for $(d)$, we apply a change of variable by setting $t= \sqrt{2}t$.\\
Moreover, we can bound the second term of \eqref{second} as follows
\begin{align*}
    \sum_{t=\tau+1}^T \exp(-\frac{Nt'C_0^2}{2}) &= \sum_{t=\tau+1}^T \exp(-\frac{N \xi^{\lfloor \log_\xi t \rfloor} C_0^2}{2}) \\
    &\le \sum_{t=\tau+1}^T \exp(-\frac{N t C_0^2}{4})\\
    &\le \int_{t=0}^{\infty} \exp(-\frac{NtC_0^2}{4})\\
    &= \frac{4}{N C_0^2}.
\end{align*}
As a result, we can bound the expected cumulative regret as follows

\begin{align*}
    R_i(T) 
	   &\le 2s_As_1\tau \notag \\ 
	   &\quad + \frac{8\sqrt{\xi}}{\sqrt{\xi} - 1} K\sqrt{\sigma^2 C_a^2 (s_0 + \frac{16 s_0 \nu C_b}{\phi_0})^2 \log^2 T + (\norm{\theta^*}_2^2 - 2 \sigma^2 \log \delta) C_a (s_0 + \frac{16 s_0 \nu C_b}{\phi_0})\log T} (\sqrt{\xi T} - 1) \notag \\
       &\quad + 2 K s_A s_1 (\frac{1-T^{1-2N}}{2N-1} + \frac{4}{N C_0^2} + \sum_{t=\tau + 1}^T \mathbb{P}((\mathcal{G}_{t,i}^{\frac{\alpha}{4 \nu C_b}})^c | \mathcal{E}_t))\\
       &\stackrel{(a)}{\le} 2s_As_1\tau \notag \\ 
	   &\quad + \frac{8\sqrt{\xi}}{\sqrt{\xi} - 1} K\sqrt{\sigma^2 C_a^2 (s_0 + \frac{16 s_0 \nu C_b}{\phi_0})^2 \log^2 T + (\norm{\theta^*}_2^2 - 2 \sigma^2 \log \delta) C_a (s_0 + \frac{16 s_0 \nu C_b}{\phi_0})\log T} (\sqrt{\xi T} - 1) \notag \\ 
       &\quad + 2 K s_A s_1 (\frac{1-T^{1-2N}}{2N-1} + \frac{4}{N C_0^2} + \sum_{t=\tau + 1}^T \exp(\log(s_0 + \frac{16 s_0 \nu C_b}{\phi_0^2}) - \frac{t' \alpha}{20 s_A \nu C_b (s_0 + (16 s_0 \nu C_b)/(\phi_0^2))}))\\
\end{align*}
where inequality $(a)$ holds true due to Lemma \ref{lem:egnval}. Now, we upper bound the last term of above equation in the following:
\begin{align*}
    &\sum_{t=\tau + 1}^{T} \exp(\log(s_0 + \frac{16\nu C_b s_0}{\phi_0^2}) - \frac{t' \alpha}{20 s_A \nu C_b (s_0 + (16\nu C_b s_0)/\phi_0^2)}) \\
    &= \sum_{t=\tau + 1}^{T} \exp(\log(s_0 + \frac{16\nu C_b s_0}{\phi_0^2}) - \frac{\xi^{\lfloor \log_\xi t \rfloor} \alpha}{20 s_A \nu C_b (s_0 + (16\nu C_b s_0)/(\phi_0^2))}) \\
    &\le \sum_{t=\tau + 1}^{T} \exp(\log(s_0 + \frac{16\nu C_b s_0}{\phi_0^2}) - \frac{t \alpha}{40 s_A \nu C_b (s_0 + (16\nu C_b s_0)/\phi_0^2)})\\
    &\le \int_{t=0}^{\infty} \exp(\log(s_0 + \frac{16\nu C_b s_0}{\phi_0^2}) - \frac{t \alpha}{40 s_A \nu C_b (s_0 + (16\nu C_b s_0)/\phi_0^2)})\\
    &\le (s_0 + \frac{16\nu C_b s_0}{\phi_0^2})^2 \frac{40 s_A \nu C_b}{\alpha}.\\
\end{align*}
That completes the proof. In summary, we have
\begin{align*}
    R_i(T) &\leq 2s_As_1\tau  \\
    &\quad + \frac{8K\sqrt{\xi}}{\sqrt{\xi} - 1} \sqrt{\sigma^2 C_a^2 (s_0 + \frac{16 s_0 \nu C_b}{\phi_0})^2 \log^2 T + (s_2^2 - 2 \sigma^2 \log \delta) C_a (s_0 + \frac{16 s_0 \nu C_b}{\phi_0})\log T} (\sqrt{\xi T} - 1) \\
	&\quad + 2 K s_A s_1 ( \frac{1-T^{1-2N}}{2N-1} + \frac{4}{N C_0^2} + (s_0 + \frac{16\nu C_b s_0}{\phi_0^2})^2 \frac{40 s_A \nu C_b}{\alpha}).
\end{align*}
\qed

\end{document}